\documentclass[twoside]{article}

\usepackage[accepted]{aistats2025}
\usepackage{verbatim}
\usepackage{graphicx}
\usepackage{amssymb,amsmath,amsfonts,mathtools}
\usepackage[colorlinks,citecolor=blue,linkcolor=black,urlcolor=black]{hyperref}
\usepackage{url}
\usepackage{listings}
\usepackage{boxedminipage}
\usepackage{enumerate}
\usepackage{tikz}
\usepackage{comment}
\usepackage{named}
\usepackage{ulem}
\usepackage{cuted}
\usepackage{algorithm}
\usepackage[font=footnotesize,labelfont=bf]{subcaption}
\usepackage[font=footnotesize,labelfont=bf]{caption}
\usetikzlibrary{calc}
\usetikzlibrary{arrows, automata}
\usetikzlibrary{shapes}
\usetikzlibrary{positioning, quotes}
\usepackage{natbib}
\usepackage{booktabs}

\newcommand{\shrink}[1]{}

\newcommand{\jin}[1]{\textcolor{blue}{[[ JT: #1]]}}
\newcommand{\rina}[1]{\textcolor{red}{[[ rina: #1]]}}
\newcommand{\annie}[1]{\textcolor{magenta}{[[ annie: #1]]}}

\usetikzlibrary{calc}
\usetikzlibrary{arrows}

\usetikzlibrary{shapes}
\usetikzlibrary{positioning, quotes}
\newcommand{\G}{\mathcal{G}}
\newcommand{\M}{\mathcal{M}}
\newcommand{\B}{\mathcal{B}}

\newcommand{\F}{\boldsymbol{F}}
\newcommand{\D}{\mathcal{D}}

\newcommand{\U}{\boldsymbol{U}} % for use in math mode
\newcommand{\V}{\boldsymbol{V}} % for use in math mode
\newcommand{\ST}{\boldsymbol{ST}} % for use in math mode
\newcommand{\E}{\boldsymbol{E}} % for use in math mode

\newcommand{\YY}{\boldsymbol{Y}} % for use in math mode
\newcommand{\XX}{\boldsymbol{X}} % for use in math mode
\newcommand{\xx}{\boldsymbol{x}} % for use in math mode

\newcommand{\PA}{P\!A}
\newtheorem{example}{Example}

\newtheorem{corollary}{Corollary}
\newtheorem{theorem}{Theorem}
\newtheorem{proof}{Proof}

\newtheorem{definition}{Definition}
\begin{document}

\twocolumn[

\aistatstitle{Graph-based Complexity for Causal Effect  by Empirical Plug-in}

%\aistatstitle{Graph-based Complexity for Causal Effect Evaluation by Plug-in}

\aistatsauthor{ Rina Dechter \And Annie Raichev\And  Alexander Ihler \And Jin Tian}

\aistatsaddress{ University of California, Irvine \And Iowa State University } 
]
%%%define query

\begin{abstract}
%%%%%%%%%%%%%%%%%%%%%%%%%%%%%%
This paper focuses on the computational complexity of computing empirical plug-in estimates for causal effect queries.
Given a causal graph and observational data, any identifiable causal query can be estimated from an expression over the observed variables, called the estimand. The estimand can then be evaluated by plugging in probabilities computed empirically from data. 
In contrast to conventional wisdom, which assumes that high dimensional probabilistic functions will lead to exponential evaluation time of the estimand. We show that computation can be done efficiently, potentially in time linear in the data size, depending on the estimand's hypergraph.
%}
In particular, we show that both the treewidth and hypertree width of the estimand's structure bound the evaluation complexity of the plug-in estimands, analogous to their role in the complexity of probabilistic inference in  graphical models. Often, the hypertree width provides a more effective bound, since the empirical distributions are sparse.

%In particular, since the empirical distribution can be characterized as being sparse, the hypertree width can more effectively bound the computational complexity of the empirical plug-in approach than the treewidth.

%%%%%%%%%%%%%%%%%%%%%%%%%%%%%%%%%%%%%%%%
  %  The paper focuses on the computational complexity of computing plug-in estimates for causal effect queries.  Given a causal graph and observational data, any identifiable causal query can be estimated from an expression over the observed variables, called the estimand. The estimand can then be evaluated by ''plugging in" the probabilities computed empirically given data generated from the observational distribution. We demonstrate that graph parameters, the treewidth and hypertree width, bound the complexity of evaluating the plug-in estimands, analagous to their role in the complexity of probabilistic inference in 
   % graphical models. In particular, since the empirical distribution can be characterized as being sparse, the hypertree width can more effectively characterizes the computational complexity of the empirical plug-in approach than the treewidth.
    
\end{abstract}

\section{Introduction}

Given a causal graph and data from the observed distribution of a Structural Causal Model (SCM), a causal effect query can be answered by a two step process: First, determine if the query is {\it identifiable}, namely if it can be answered uniquely given the graph and observational data, and if so generate an estimand for the query. The estimand is an algebraic expression over probabilistic functions that involve observed variables only.
Second, the estimand is evaluated using data from the observational distribution. A straightforward approach for evaluation is the empirical ``plug-in" method, which simply replaces the probabilistic functions in the estimand with the empirical probabilities from the data.

%\jin{One could use other estimators for conditional probabilities, e.g. Laplace, Bayesian, ML estimators. To my understanding, any method that plugs in the estimated CPDs into the estimand is called a plug-in method (is this the case?). (1) We need to clarify/restrict the scope of the paper to the particular plug-in method that uses empirical CPTs - perhaps give it a name, "empirical plug-in"? (2) Alternatively, should we talk about the complexity of other plug-ins that have no zero entries in CPT estimation too, and compare the performance of empirical plug-in against other plug-ins?  }

%\jin{Discuss motivation for investigating complexity of plug-in as suggested by Alex. Use a concrete example?}
However, the estimand expression often involves high dimensional conditional probability functions 
%\jin{
and marginalization over a large number of variables (e.g., see Eq.~\eqref{eq:cone}). %}. %, which seems to suggest that the expression is difficult or impossible to evaluate. 
A common assumption is that, for discrete models, the computation required is at least the size of the largest table, and thus exponential in the number of arguments of the largest term in the expression.  
This assumption suggests that the estimand expression is difficult or impossible to evaluate in high-dimensional settings.
%Easy to create examples where this is the entire model size. 
%Yet, taking into account the data size, the initial sizes of the empirical probability tables are bounded so estimand evaluation can also be bounded by the data-size, in some cases linearly, regardless of the functions dimension. \jin{however, this is a misunderstanding too?}
However, taking into account the data size, the initial sizes of the empirical probability tables are bounded, regardless of the functions' dimension. So, the estimand evaluation can also be bounded in terms of the data size, in some cases linearly. 

%ATI: maybe rewrite?  something like
 %(0) While a number of approaches have been proposed for evaluating the estimand expression [cite], perhaps the most straightforward is the plug-in estimate...
 %(1) [why is estimand hard] Unfortunately,
% the estimand expression often involves very high dimensional (conditional) probability functions, making it difficult or impossible to evaluate. A common assumption is that, for discrete models, the computation required is at least the size of the largest table, or exponential in the number of arguments of the largest term in the expression.  Easy to create examples where this is the entire model size.
% (2) On the other hand, 
%
%
%
%
%
%
%

Our paper focuses on the complexity of computing plug-in estimates. We explore the impact of both the functions' dimension and the data size on the complexity of evaluation.
We show that well-known graph parameters such as {\it tree-width} and {\it hypertree width}, which play a central role in the complexity of probabilistic inference, play a similar role in plug-in estimand evaluation.  It is well known that probabilistic inference is exponential in the tree-width \citep{Dechter:2003aa,DBLP:series/synthesis/2013Dechter}. However, when a graphical model's functions are sparse (e.g., have many zeros), the {\it hypertree width} can provide a tighter exponential bound than treewidth. \citep{DBLP:journals/ai/GottlobLS00,DBLP:journals/ai/KaskDLD05,DBLP:conf/uai/OttenD08}. %It was shown %in %\citep{DBLP:journals/ai/GottlobLS00,DBLP:journals/ai/KaskDLD05}, 
%the hypertree width may provide a more effective characterization of tractablility than treewidth when the model's functions are sparse. 
We build on these results to show that the tree-width and hypertree width play a similar role in the complexity of plug-in estimand evaluation. Since, the empirical probabilities are inherently sparse, the hypertree width is often far more informative than the tree-width for this task. 
We associate an estimand expression with a subexpression hierarchy and show how the hypertree widths of subexpressions  additively determine complexity bounds on estimand evaluation. 
Our bounds help
 characterize the computational feasibility of the 
 empirical plug-in scheme, 
 %and the choice of its parameters (e.g., number of samples) from a computational point of view
 % AI: took this out since # of samples is not usually in our control
 establishing it as a simple and practical baseline for causal effect estimation.
 Moreover, since  a causal query can have many candidate estimands, tree-width and hyperwidth-based bounds can be used as one metric to selecting among different estimands. 
 %Finally, we applied our new plug-in hypertree evaluation \annie{you said we didnt run this technically}evaluation algorithm over several benchmarks and illustrated the effectiveness of the hypertree width in capturing the actual running time and memory in empirical plug-in estimation. 
 Finally, we illustrate the effectiveness of the hypertreewidth in capturing the actual time and memory bounds of empirical plug-in estimation. 
 
%\sout{Finally, our results establish the empirical plug-in method as a computationally practical baseline for causal effect estimation. }?
%\shrink{\rina{I am not fully sure what this sentence says.}\annie{in response to this comment by jin, I took this to mean that since the empirical plug is computationally tractable, we can claim it to be a beneficial baseline}\jin{Can we claim as contribution establishing plug-in as a baseline for comparing with more sophisticated causal effect estimation methods?}}

%\paragraph{Significance.} Given an estimand and data-size we can characterize the feasibility of computation by the plug-in scheme using our bound. Moreover, since  a causal query can have many candidate estimands, the tree-width and hyperwidth-based bounds can be used as one metric in selecting among different estimands.

%\paragraph{Contributions.} The paper shows how established complexity results on inference in the presence of sparse functions can be extended to estimand evaluation for causal effect queries. It associates an estimand with a hypergraph hierarchy and shows how the hypertree widths of hypergraphs in different levels of this hierarchy additively determine a complexity bound on the estimand evaluation. This establishes the empirical plug-in method as a computationally practical baseline for causal effect estimation

\section{Background}
\shrink{
%\paragraph{Notations.} 
We use capital letters $X$ to
represent variables, and lower case $x$ for their values.
Bold uppercase ${\bf  X}$ %${\XX}$ 
denotes a collection of variables, with
$|{\bf X}|$ its cardinality and ${\bf D}({\bf X})$ their joint domain, while $\bf x$ indicates an assignment in that joint domain.
We denote by $P({\bf X})$ the joint probability 
distribution over ${\bf X}$ and $P({\bf X} = {\bf x})$ the probability of ${\XX}$ taking configuration $\xx$ under this distribution, which we abbreviate  $P({\bf x})$.
}
We begin with some useful definitions and notation.
\begin{definition}[Structural Causal Model]
A structural causal model (SCM) \citep{pearlbook} is a $4$-tuple $\M = \langle \U, \V, \F, P(\U) \rangle$ where:
(1) $\U
= \{U_1, U_2, ..., U_k\}$ is a set of exogenous (latent) variables whose values are affected by outside factors;
   (2) $\V =
\{V_1, V_2, ..., V_n\}$ are endogenous, observable
   variables; 
   %whose values are determined by other variables in the model;
(3) $\F = \{f_{i}: V_{i} \in \V\}$ is a set of functions $f_{i}$ where each $f_i$ determines the value $v_{i}$ 
of $V_{i}$ as a function of $V_{i}$'s causal parents $\PA_{i} \subseteq \U\cup(\V\setminus V_{i})$ 
so that %$f_{i} : \D(\PA_{i}) \rightarrow \D(V_{i})$ and  \jin{$\D$ not defined}
$v_{i} \leftarrow f_{i}(pa_{i})$; 
(4) $P(\U)$ is a probability distribution over the
latent variables. The latent variables are assumed to be mutually independent. i.e., $P(\U) = \prod_{U \in \U} P(U)$. 

\end{definition}
%\subsection{Causal Effect Evaluation}

\paragraph{Causal diagrams.} The {\it causal diagram} of an SCM $\M$ is a directed graph $\G = \langle \V \! \cup \U, E \rangle$, where each node represents a variable, and  
there is an arc in $E$ from a node representing $X$ to a node representing $Y$ iff  $X$ is a parent of $Y$. We assume semi-Markovian SCMs in which latent variables  connect to at most two observable variables \citep{jinthesis}.
\shrink{
An SCM whose latent variables connect to at most a single observable variable is referred to as {\it Markovian}, and one whose latent variables connect to at most two observable variables is called {\it Semi-Markovian}. It is known that any SCM can be transformed into an equivalent Semi-Markovian one such that answers to causal queries are preserved \citep{jinthesis}.
}
Here it is common to 
%use a simplified causal diagram called an \textit {Acyclic Directed Mixed Graph} (ADMG), 
%which 
omit latent variables having a single child, and 
replace any latent variable with a bidirectional dashed arc between the children 
(see Figure
\ref{fig:chain_7}, \ref{fig:cone_cloud}, \ref{fig:diamond}). 

An SCM $\M$ induces a {\bf Causal Bayesian Network (CBN)} $\B = \langle \G, \mathcal{P} \rangle$ specified by $\M$'s causal diagram $\G = \langle \V \! \cup \U, E \rangle$ along with its associated conditional probability distributions $\mathcal{P} = \{P(V_{i}|\PA_{i}),  P(U_j)\}$. 
The distribution $P(\V, \U)$ factors according to the causal diagram:
\begin{align}\label{eq-cbn}
 P(\V, \U) =  \prod_{V_i \in \V} P(V_{i}|\PA_{i}) \cdot \prod_{U_j \in \U} P(U_j).
\end{align}
The \textit {observational distribution},  $P(\V)$, is given by 
\begin{equation}\label{eq-obs}
P(\V) = \sum_{\U}  P(\V, \U).  
\end{equation}
\vspace{-3mm}
%\shrink{ 
\paragraph{Causal effect and the truncation formula.} 
An external intervention forcing variables $\XX$ to take on value $\xx$, called $do(\XX=\xx)$, is modeled by replacing the mechanism for each $X \in \XX$  with the function $X=x$. Formally,
\vspace{-2mm}
\begin{equation} \label{eq-intervene}
P(\V\setminus\XX,\U \mid do(\XX=\xx) ) = \prod_{V_j \notin {\XX}} P(V_{j}|\PA_{j}) \cdot P(\U) \bigg|_{\XX=\xx} 
\end{equation}
Namely, it is obtained from Eq.~\eqref{eq-cbn} by truncating the factors corresponding to the variables in $\XX$ and setting $\XX=\xx$. 
%i.e., it is derived from the original causal model by removing the function for $X$ and setting $X=x$. 
The effect of $do(\XX)$ on a variable $Y$, denoted $P(Y| do(\XX))$, is defined by marginalizing all the variables other than $Y$. 
%}

%\paragraph{Causal queries.} 
The standard formulation of causal inference assumes that we only have access to the causal graph $G$ and the observational distribution  $P(\V)$ (or a sample from it). The identifiability task is to determine if the query can be {\it uniquely} answered from $G$ and $P(\V)$. This occurs if the answer is unique for any full model that is consistent with the graph and $P(\V)$ \citep{pearlbook}. In such cases an estimand expression in terms of $P(\V)$  can be generated and evaluated.

\shrink{ 
\paragraph{Causal queries.} 
While we normally have no access to the full structural causal model $\M$ and cannot evaluate these expressions directly, it 
is sometimes possible to evaluate the effect of an intervention given only the observational distribution $P(\V)$,
specifically, if the answer is unique for any full model that is consistent with the graph and $P(\V)$ \citep{pearlbook}. More formally:
\begin{definition}[Identifiability]
 Given a causal diagram $\G$, the causal effect query $P({\YY} \mid do({\XX}))$ is {\bf identifiable} if any two SCMs consistent with $\G$ that agree on the observational distribution $P(\V)$ 
 also agree on $P(\textbf{Y} \mid do(\textbf{X}))$. 
\end{definition}
}
%\annie{Looks like this is a redundant defintion of the one above?} \rina{here we explicitly say we have a causal graph and data. It is not stated like that above.}
\begin{definition}[Causal-effect query] 
Given a causal diagram $\G=\langle \V \! \cup  \U, E \rangle$, data samples from the observational distribution $P(\V)$, and an identifiable query $P({\YY} \mid do({\XX}))$,
the task is to compute the value of $P({\YY} \mid do({\XX}))$. 
\end{definition}

\paragraph{Estimand-based approaches.}
The now-standard methodology for answering causal-effect queries is to break the task into two steps.
The first is the \emph{identifiability step}: given a causal diagram and a query $P(\YY \mid do(\XX))$, determine if the query is identifiable and if so, generate an \emph{estimand}, or algebraic expression 
in terms of the observational distribution $P(\V)$ that answers the query.
A complete polynomial algorithm called ID has been developed for this task \citep{jinthesis,shpitser2006identification}.
The second step is \emph{estimation}: use samples from the observational distribution $P(\V)$ to estimate the value of the estimand expression. 
A number of approaches have been applied to estimation.
%in the second step.
A simple approach, called the \emph{plug-in estimator}, replaces each term in the estimand with its empirical probability value in the observed data.
More sophisticated approaches have been developed recently \citep{DBLP:conf/aaai/Jung0B20,DBLP:conf/nips/Jung0B20, DBLP:journals/corr/abs-2408-14101}. %\jin{cite our ECAI paper? maybe in the introduction}

A {\it CBN} and {\it SCM} belong to the class of probabilistic graphical models. A graphical model is defined by a collection of functions 
over subsets of variables: 
%each defined over a subset of variables, as defined next. 
%In Bayesian networks the functions are CPTs. %In SCM the functions are the deterministic functions and the stochastic functions over the latent variables.
%
 \begin{definition}
 \label{def3}
     A {\bf probabilistic graphical model } is a triplet $\M=\langle \XX,
{\bf D}, \F  \rangle$ 
%\jin{can we use a different notation $\M$ to distinguish with SCM $\M$?} 
where ${\bf X} = \{X_1, \ldots, X_n\}$ is a set of variables with finite domains
 $D =
\{D_1, \ldots, D_n\}$, and $\F = \{f_1, \ldots, f_r\}$ is a set of discrete real-valued functions, each
  defined over a subset of variables $S_i \subseteq {\bf X}$, called its scope and
  denoted by $scope(f_i)$. The product of all the function in the model defines a probability distribution (Bayesian networks or Markov network) and a variety of reasoning problems can be defined using marginalization operator such as $sum_X, max_X$. The {\bf primal graph} of a graphical model is an undirected graph whose nodes are the variables and edges connect every two nodes that appear in the scope of a single function. The {\bf dual graph} is a graph whose nodes are the scopes $\{S_1,...,S_r\}$  of the functions, with edges connecting any two nodes whose scopes share variables. The {\bf hypergraph} of a graphical model has the variables as its nodes and the hyperedges are again the scopes of the functions.

\shrink{
  %%%%%%%%%%%%%%%%%%%%%%%%%%%%%%%%%%
  The graphical model defines a global function obtained by a combination operator like product or summation or multiplication, and a variety of reasoning problems can be defined using a marginalization operator such as $sum_X, max_X$. 
  %The {\bf primal graph} of a graphical model is an undirected graph whose nodes are the variables and edges connect every two nodes that appear in the scope of a single function.
  \rina{maybe revise?} \jin{How is this definition anything to do with "graph"? I'd think the name "graphical model" is rather misleading. The definition of global function is quite ambiguous. Is the intention here to define an inference task/query? I think a query task should be defined here to ground the meaning of "inference".}
  %we obtain a reasoning problem.
  }
  \end{definition}
A Bayesian network over ${\bf X} = \{X_1, \ldots, X_n\}$ is a probabilistic graphical model with $f_i(X_i, PA_{X_i}) = P(X_i|PA_{X_i})$ for $i=1, \ldots, n$.
  \begin{example} \upshape
 Consider the belief
network 
in Figure \ref{fig:belief-network}. It contains variables
$A,B,C,D,F,G$ and  functions $f(A,B)$, $f(A,C)$, $f(B,C,F)$,
$f(A,B,D)$, $f(F,G)$.
%\sout{which are conditional probability tables %(CPT) 
%between a child node and its parents. For example %$f(B,C,F) =
%P(F|B,C)$. } 
The primal graph of the model is depicted in Figure \ref{fig:moral_graph}. The dual graph and hypergraph are not explicitly illustrated here, for space reasons.

%\jin{What's the point of this example? need revising.} \rina{to illustrate a graphical model in general}
\end{example}

\shrink{

\begin{figure}[t]
    \begin{subfigure}[h]{.5\columnwidth}
    \centering
         \resizebox{\textwidth}{!}{\includegraphics{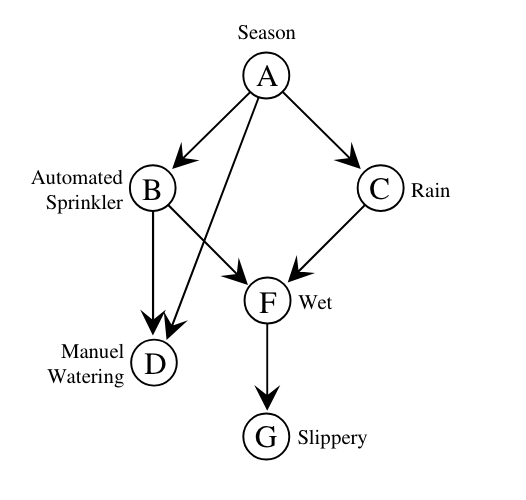}} 
        \captionsetup{justification=centering}
         \caption{Belief network\\$P(g,f,d,c,b,a)$}
         \label{fig:belief-network}%
      \end{subfigure}\hfill 
    \begin{subfigure}[h]{.5\columnwidth}
        \centering
         \resizebox{.7\textwidth}{!}{\includegraphics{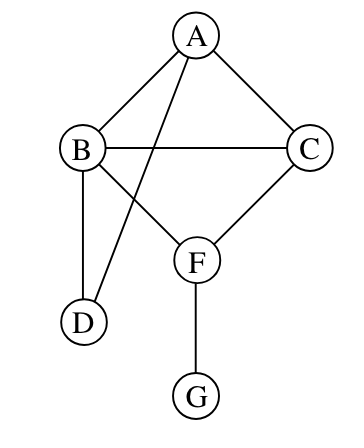}} 
         \caption{Primal graph}
         \label{fig:moral_graph}%
      \end{subfigure}\\
    \begin{subfigure}[b]{.5\columnwidth}
         \resizebox{.9\linewidth}{!}{\begin{tikzpicture}[-,>=stealth',shorten >=1pt,auto,node distance=1.5cm,
  main node/.style={ellipse,draw,font=\sffamily\small},
    every label/.style={font=\sffamily\small}]
  
            \node[main node, label=above:{$f(G,F)$}] (1) {$F,G$};  
            \node[main node, label=above left:{$f(B,C,F)$}] (2)[below left of=1] { $B,C,F$};
            \node[main node, label=above right:{$f(A,B), f(A,C)$}] (3)[below right of=2]  { $A,B,C$};
            \node[main node, label=above left:{$f(A,B,D)$}] (4)[below left of=3]{$A,B,D$};

                \draw[thick] (1) edge  (2);
               \draw[thick] (2) edge  (3);
                \draw[thick] (3) edge  (4);

\end{tikzpicture}} 
         \caption{Tree decomposition 1 }
         \label{fig:tree_decomp1}%
      \end{subfigure}\hfill 
        \begin{subfigure}[b]{.5\columnwidth}
         \resizebox{.9\linewidth}{!}{\begin{tikzpicture}[-,>=stealth',shorten >=1pt,auto,node distance=2cm,
  main node/.style={ellipse,draw,font=\sffamily\small},
    every label/.style={font=\sffamily\small}]
        
            \node[main node, label=above:{$f(A,B), f(A,C), f(A,B,D) \}$}] (1) {$A,B,C,D$};  
            \node[main node, label=above left:{$f(B,C,F)$}] (2)[below left of=1] { $B,C,F$};
            \node[main node, label=above right:{$f(F,G) $}] (3)[below right of=2] { $F,G$};

                \draw[thick] (1) edge  (2);
               \draw[thick] (2) edge  (3);

        \end{tikzpicture}} 
         \caption{Tree decomposition 2}
         \label{fig:tree_decomp2}%
      \end{subfigure}
          \caption{Graphical models and their tree decompositions}
      \label{fig:bayesian-ex}
      \vspace{-4mm}
\end{figure}
}
\begin{figure}[t]
    \begin{subfigure}[t]{.45\columnwidth}
    \centering
         \resizebox{\textwidth}{!}{\includegraphics{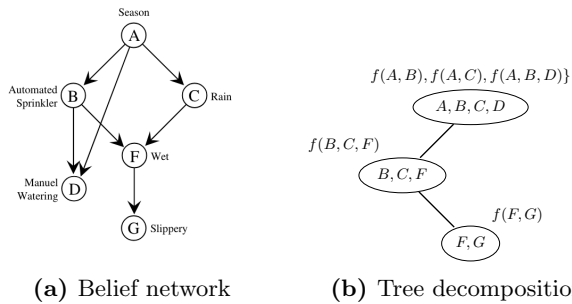}} 
        \captionsetup{justification=centering}
         \caption{Belief network}
         \label{fig:belief-network}%
      \end{subfigure}\hfill 
        \begin{subfigure}[t]{.5\columnwidth}
         \resizebox{.9\linewidth}{!}{\begin{tikzpicture}[-,>=stealth',shorten >=1pt,auto,node distance=2cm,
  main node/.style={ellipse,draw,font=\sffamily\small},
    every label/.style={font=\sffamily\small}]
        
            \node[main node, label=above:{$f(A,B), f(A,C), f(A,B,D) \}$}] (1) {$A,B,C,D$};  
            \node[main node, label=above left:{$f(B,C,F)$}] (2)[below left of=1] { $B,C,F$};
            \node[main node, label=above right:{$f(F,G) $}] (3)[below right of=2] { $F,G$};

                \draw[thick] (1) edge  (2);
               \draw[thick] (2) edge  (3);

        \end{tikzpicture}} 
         \caption{Tree decomposition}
         \label{fig:tree_decomp2}%
      \end{subfigure}
          \caption{Graphical Model And Its Tree Decomposition}
      \label{fig:bayesian-ex}
      \vspace{-4mm}
\end{figure}

%%%%%%%%%%%%%%%%%%%%%%%%%%%%%%
%%%%%%%%%%removing graph concepts starting %%%%%%%%%%%%%%%here
%%%%%%%%%%%%%%%%%%%%%%%%%%%%%%%%%%%%%%%%%%

%\shrink{
%\shrink{

\subsection{Graph concepts}
\label{sec-prelim}
%\jin{Do we really need this section? It looks to me these concepts of hypergraph, primal/dual graph, hypertree can all be avoided since tree/hypertree decomposition is actually defined over a graphical model.}

Graphs are an integral components of graphical models, so we next
define well-known concepts over graphs and hypergraphs and their relation to graphical models. 
For more details see  \citet{DBLP:journals/ai/KaskDLD05,DBLP:series/synthesis/2013Dechter, DBLP:books/cu/p/GottlobGS14}.

\begin{definition}[hypergraph]
The hypergraph of a graphical model $\M = <{\bf X}, {\bf D}, \F >$ is a pair $H=(\V,\ST)$ where $\V = {\bf X}$ and 
%\sout{$\ST=\{S_1,...,S_t\}$
is a set of subsets of $\V$,  called hyper-edges, %which represent 
%the scopes of the functions ($S_i = scope(f_i)$).} $\ST=\{S_1,...,S_r\}$ is the set of the scopes of the functions  in $\F$ ($S_i = scope(f_i)$), called hyper-edges.}%\jin{Does $\ST$ contains $scope(f_i)$ for all $f_i \in F$ or not? Can a model $\M$ have multiple hypergraphs or just one? The definition needs revision if a model is associated with a unique hyergraph.} 
%\annie{ added defintion for \bf{primal graph}
%} 
The {\bf primal graph} of a hypergraph $H=(\V,\ST)$ is an undirected
graph $G = (\V,\E)$ such that there is an edge $(u,v) \in \E$ for any two vertices $u,v \in \V$ that appear in the same hyperedge. The {\bf dual graph} of a hypergraph is a graph where each hyperedge is a node, and two nodes are connected by an edge if their nodes have non-empty intersection.
%The edges are labeled by the intersection sets.
\end{definition}
%}

\begin{definition}[hypertree]
A hypergraph is a {\bf hypertree}, also called {\bf acyclic}, 
%hypergraph}, 
if %and only if 
its dual graph has an edge subgraph
that is a tree (called a join-tree) satisfying that all its nodes 
%in the subgraph that
that contain a common variable form a connected subgraph.  This condition is also known as the ``running intersection property'' or the ``connectedness property''. A {\bf join-graph} is an edge subgraph of the dual graph that satisfies the conectedness property. If the hypergraph of a graphical model is a hypertree, 
the graphical model is called {\bf acyclic}.

%\jin{I have trouble understanding this definition.} \rina{I tried a different definition. I think this one is better.} \jin{I can't make sense of this definition. Isn't the condition "such that ..." automatically satisfied by the definition of dual graph? I guess "such that all the nodes in the dual graph ... " should be "such that all the nodes in this tree ..."?}\rina{The dual graph always satisfies the condition. But if you remove arcs from the dual graph 2 nodes that share a variable may not be connected by an edge anymore. This is still allowed if there is an alternative path  s.t all the nodes (which are sets of variables) on the path includes that variable. If you can remove edges only if the condition is satisfied and get to a tree, the hypergraph is a hypertree... This is a standard property in a tree-decomposition.}
\end{definition}

\shrink{
\begin{definition}[hypertree] \rina{is this definition better?}
A hypergraph $H=(V,S)$ is a {\bf hypertree}, also called {\bf acyclic
hypergraph}, if and only there exists a tree structure spanning all  its edges as nodes such that 
 for any $v \in V$ all the edges (sets) that contain $v$ form a connected subgraph.
%\jin{I have trouble understanding this definition.}
\end{definition}
}

\shrink{
\begin{definition}[induced width]
An {\bf ordered graph} is a pair $(G,d)$  denoted $G_d$ where $G$
is an undirected graph, and $d = (X_1,...,X_n)$ is an ordering of
the vertices. The {\it width of a vertex} in an ordered graph is
the number of its earlier neighbors. The {\it width of an ordered
graph}, $w(G_d)$, is the maximum width of all its vertices. The
{\bf induced width of an ordered graph}, $w^*(G_d)$, is the width
of the induced ordered  graph, denoted $G^*_d$, obtained by
processing the vertices recursively, from last to first; when
vertex $X$ is processed, all its earlier neighbors are connected.
The {\it induced width of a graph}, $w^*(G)$, is the minimal
induced width over all its orderings \citep{dechter88}.
\end{definition}
}

%%%%%%%%%%%%%%%%%%%%%%%%%%%%%%%%%%%%%
%%%%%removing graph concepts
%%%%%%%%%%%%%%%%%%%%%%%%%%%%%%%%%%%

%}

%\label{sec-CTE}

\subsection{Tree and Hypertree Decompositions}
\label{sec-tree-decomposition}

Tree decomposition schemes have been widely used  for constraint
processing and probabilistic reasoning.
The most popular
variants are join-tree (also known as  junction-tree) algorithms, which include variable elimination schemes \citep{Dechter:2003aa,DBLP:journals/ai/GottlobLS00,DBLP:series/synthesis/2013Dechter}. The methods
vary somewhat in their graph definitions as well as the way the
tree decomposition is processed. However,
all involve a decomposition of a hypergraph of a graphical model into a hypertree:
%, as will be defined next.

\begin{definition}[tree \& hypertree decompositions] \citep{DBLP:journals/ai/KaskDLD05,DBLP:journals/ai/GottlobLS00}
%Let $P = \langle {\cal R} \Downarrow, \{Z_i\} \rangle$  be a reasoning problem
%over a graphical model $\langle X, D, F, \bigotimes \rangle$. 
A {\bf
  tree-decomposition} of a graphical model $\M = \langle \XX,
{\bf D}, \F  \rangle$
  is a triplet $\langle T, \chi,
\psi \rangle$, where $T = (\V, E)$ is a tree, and $\chi$ and $\psi$ are
labeling functions that associate with each vertex $v \in \V$ two
sets, $\chi(v) \subseteq \XX$ and $\psi(v) \subseteq \F$, that satisfy the following conditions:
\begin{enumerate}
\item For each function $f_i \in \F$, there is {\bf exactly one}
vertex $v \in \V$ such that $f_i \in \psi(v)$.
\item If $f_i \in
\psi(v)$, then $scope(f_i) \subseteq \chi(v)$.
\item For each
variable $X_i \in \XX$, the set $\{v \in \V | X_i \in \chi(v)\}$
induces a connected subtree of $T$. This is also called the
running intersection or the connectedness property. %\jin{This sentence repeats that in Definition 5.} \rina{Good point: maybe definition 5 is not needed.}
\end{enumerate}

The {\bf treewidth} $w$  of a tree-decomposition $\tau = \langle T, \chi,
\psi \rangle$ is ${\max_{v \in V}} |\chi(v)|-1$. %${max \atop {v \in V}} |\chi(v)|-1$. 
%\jin{
The treewidth of a graphical model is the minimum treewidth over all its tree-decompositions.
%...}

The tree-decomopsition $\tau$ is also called a {\bf hypertree decomposition} of the graphical model if it satisfies the additional condition that the variables in each node, also called a cluster, are covered by the arguments of the functions in the cluster. Formally:
\begin{enumerate}
%\setcounter{enumi}{3}
%\item For each $v \in V$, $\chi(v) \subseteq scope(\psi(v))$. 
\vspace{-2mm}
\item[4.] For each $v \in \V$, $\chi(v) \subseteq \cup_{f_j \in \psi(v)}  scope(f_j)$. 
\vspace{-2mm}

\end{enumerate}
In this case the  {\bf hypertree width} of $\tau$ is 
%\annie{should use $\mathrm{hw}$ for hw} \rina{italic} 
$hw = max_v
|\psi(v)|$. The hypertree-width of a graphical model is the minimum
hypertree width over all possible hypertree decompositions 
 of the graphical model. 
 %\jin{``of the graphical model associated with the hypergraph'', since technically, the tree-decompositions of a hypergraph is not defined} \rina{Yes. though it can be defined on the hypergraph directly.}.
\label{CT-def}
\end{definition}

Finding tree or hypertree decompositions having the minimal treewidth or hyperwidth is known to be NP-hard. Although optimizing the hypertree width is harder in the sense that verifying that a graph has a hypertree width smaller than $w$ is not polynomial %\jin{proved to be non polynomial or no known polynomial algorithm available?} \rina{I think it was proved be hard but I will have to go back and check.} , 
like it is for treewidth \citep{DBLP:journals/jea/GottlobS08,DBLP:journals/ai/SchidlerS23}. Heuristic algorithms are employed in practice for both parameters; most of these center on finding a good ordering of the variables that leads to a small treewidth or hyperwidth. For details see \citet{Dechter:2003aa,DBLP:conf/aaai/KaskGOD11,DBLP:books/cu/p/GottlobGS14,
%DBLP:conf/uai/OttenD08,
DBLP:journals/corr/abs-1207-4109}.

%\annie{should we say something about these algorithms, like %name them} \rina{Yes. Good suggestion}
%\jin{Provide algorithm names? Our algorithm will need to run such an algorithm in the middle.} \rina{I don't think we need to name it. This is such a well known problem and there are many algorithms. But if we will describe the algorithm in detail we will mention it.}

\shrink{ 
\begin{definition} [treewidth, separator]
The {\it treewidth} $w$  of a tree-decomposition $\langle T, \chi,
\psi \rangle$ is ${max \atop {v \in V}} |\chi(v)|-1$. Given two adjacent
vertices $u$ and $v$ of a tree-decomposition, a separator of $u$
and $v$ is defined as $sep(u,v) = \chi(u) \cap \chi(v)$.
\end{definition}
}

\shrink{ Notice that it may be that $sep(u,v)=\chi(u)$ (that is,
all variables in vertex $u$ belong to an adjacent vertex $v$). In
this case the size of the tree-decomposition can be reduced by
merging vertex $u$ into $v$ without increasing the tree-width of
the tree-decomposition. A tree-decomposition is {\it minimal} if
$sep(u,v) \subset \chi(u)$ and $sep(u,v) \subset \chi(v)$.}

\shrink{
\begin{figure}
\centerline{\psfig{figure=../kalev/CTDecomp/fig6.eps,width=2.5in}}
%\centerline{\psfig{figure=../../fabio/fig6.eps,width=3in}}
\caption{Several tree-decompositions of the same belief network}
\label{fig-super}
\end{figure}
}

\begin{example} (Taken from \citet{DBLP:journals/ai/KaskDLD05})
\upshape Consider the belief network in Figure
\ref{fig:belief-network} with hypertree decomposition in Figure~\ref{fig:tree_decomp2}.
%with primal graph in Figure \ref{fig:moral_graph}.
%Any of the trees in Figure~\ref{fig:bayesian-ex}
%Both of the trees in Figures~\ref{fig:tree_decomp1} and~\ref{fig:tree_decomp2}
%The trees in Figures~\ref{fig:tree_decomp1} and~\ref{fig:tree_decomp2}
Here the labeling
$\chi$ is the set of variables in each node. The functions
can be assigned to nodes whose $\chi$ variables contain their
scopes. For example, 
%in the tree-decomposition of Figure \ref{fig:tree_decomp2}, %shows a
%tree-decomposition with two vertices, and labelling $\chi(1) =
%\{G, F\}$ and $\chi(2) = \{A, B, C, D, F\}$.
any function with scope $\{G\}$ must be placed in vertex $(GF)$ because
it is the only vertex that contains variable $G$. 
%\sout{Any function with scope $\{A,  D\}$ or its subset must be placedin vertex $(ABCD)$}.
Alternatively, any function with scope $\{F\}$ can be placed
either in vertex $(FG)$ or $(BCF)$.

\end{example}
%}

%\subsection{Function Specification}
%Here we elaborate on the data structures for specifying discrete functions.

\paragraph{Relational vs.~tabular representation.} The common brute-force method
for specifying functions as a table is exponential in their
scopes, where the base of the exponent is the maximum domain size $k$.
%Relations, or clauses, can be expressed as functions that associate a value of ''0" or ''1" with each tuple, in the cartesian product of their domains, depending on whether or not the tuple is in the relation (or satisfies a clause).
%\paragraph{Relational representation.} 
We can also assume a
relational specification where all the tuples that are mapped to a
value ``0" are removed (e.g., zero probabilities) from the table. In this case, the
function is specified by $t$ tuples of length $l$ for functions
defined over scopes of size $l$ variables. The value $t$ is often referred to as
the \textit{tightness} of the function representation. Clearly, the size of the
specification can be far smaller than the worst-case of $O(k^l)$. In this paper we will assume that functions are always specified relationally.
\begin{definition}[tightness]
    The tightness of a function $f$, $t(f)$, is the number of non-zero  configuration in its domain.
    The tightness of a graphical model having the set of functions $F$ is $t = max_{f \in F} t(f)$.
\end{definition}

%\jin{I think this section should be expanded and clarified as it looks to me this representation differences are a key leading to different complexity. I think we should formally define the "tightness" $t$ of a hypertree decomposition.} \rina{the tightness is a function of the graphical model only. Not the decomposition. We should definitely define it.}

%\annie{ also dont think this paragraph is necessary either} \rina{It may be good to put this in the larger context.} Other specifications of functions can be used such as decision
%trees or decision graphs, CNF representations, causal
%independence specifications as well as algebraic representations,
%linear or non-linear. In each of these cases the size of the
%specification can be far smaller than the worst-case of $O(k^r)$.
%}

%%%%%%%%%%%%%%%%%%%%%%%%%%%%%%
\subsection{Complexity of Tree Decomposition}

\shrink{
\begin{small}
\begin{figure}
\fbox{\parbox{3in}{
{\bf Algorithm cluster-tree elimination (CTE)}\\
%\bigskip
{\bf Input:} A tree decomposition $<T, \chi, \psi>$ for a problem
$P = <X, D, F, \bigotimes, \Downarrow>$ \\
%$F= \{f_1,...,f_r\}$.\\
%{\bf Output:} An augmented tree whose vertices are clusters
%containing the original functions as well as messages received
%from neighbors.
%A solution computed from the augmented clusters.\\
%\bigskip
{\bf Compute messages:} \\
%Let $m$ be the number of edges in the cluster tree.
{\bf For} every edge $(u,v)$ in the  tree, do
\begin{itemize}
\item Let $m_{(u,v)}$ denote the message sent by vertex $u$ to
vertex $v$. Let $cluster(u) = \psi(u) \cup \{m_{(i,u)} | (i,u) \in
T\}$. If vertex $u$ has received messages from all adjacent
vertices other than $v$, then compute and send to $v$,
\[
m_{(u,v)}  = \Downarrow_{sep(u,v)} (\bigotimes_{f \in cluster(u),
f \neq m_{(v,u)}} f) \label{eq-cte}
\]
\end{itemize}
%{\bf Endfor} \\
%Note: functions whose scope does not contain elimination variables
%do not need to be processed, and can instead be directly
%passed on to the receiving vertex. \\
%\bigskip
{\bf Return:} A tree-decomposition augmented with messages. %, and
%for every $v \in T$ and every $Z_i \subseteq \chi(v)$, compute
%$\Downarrow_{Z_i} \bigotimes_{f \in cluster(v)} f$.
} }
\caption{Algorithm Cluster-Tree Elimination (CTE)} \label{alg-CTE}
\end{figure}
\end{small}
}

Once a (hyper) tree-decomposition of a graphical model is generated, any sum-product query (e.g., $P({\bf X}|{\bf Y=y})$ where {\bf X} and {\bf Y} are subsets of variables) can be answered by a message passing
algorithm
%between the nodes in the tree.
%Algorithm {\it Cluster-Tree Elimination (CTE)}
%in Figure \ref{alg-CTE}
%is a message-passing algorithm,
where each vertex of the tree sends a function to each of its neighbors.
%\annie{I think below should be cut since we explain it already for CTE, we can then end this paragraph saying We use the message passing algoithm CTE, described below:}
%\jin{Unclear what "process" means here. What is the task/query?}
\shrink{
If the tree contains $m$ edges, then a total of $2m$ messages will
be sent as follows: for
each neighbor $v$, node $u$ takes all the functions in $\psi(u)$
and all the messages received by $u$  from all adjacent nodes other than $v$, 
generates their {\it product} function which is {\it marginalized}
over the separator; namely the set of variables in the intersection
%\jin{"separator" never defined} 
between $u$ and $v$, and sends the result  to $v$. The algorithm can be
applied with any style of function specification over which the product 
and marginalization (e.g., summation) are well defined.
%For a discussion of various styles of algorithms
%such as {\it join-tree clustering (JTC)} and the more %generic version of
%{\it Cluster Tree Elimination (CTE)} see \citet{DBLP:journals/ai/KaskDLD05}.
}
We will use  Cluster-tree Elimination ({\it CTE}) \citep{DBLP:journals/ai/KaskDLD05} as a generic name for a message passing algorithm over a tree-decomposition.
%For completeness and clarity some of its details are described next, and are followed by the corresponding complexity bounds.

\paragraph{Algorithm CTE.} %\rina{maybe remove?} \jin{The CTE algorithm should be described before Theorems 1, 2 to ground them.} 
The following 3 steps define the basics of the $CTE$ algorithm \citep{DBLP:journals/ai/KaskDLD05}.
%The basics of algorithm $CTE$ 
%\annie{this sentence doesnt make grammatical  sense}\sout{which involve the message a node in the tree decomposition sends to a neighbor. It is shown in order to communicate some of the inner computation involved.}
Given a
hypertree decomposition, each node $u$ has to send a single
message to each neighbor $v$. We can compute $m_{(u,v)}$ as
follows: %{\bf Computing EQ. \ref{eq-cte}:}
\vspace{-2mm}
\begin{enumerate}
\item Combine all functions $\psi(u)$ in node $u$ yielding function
$h(u) = \prod_{f \in \psi(u)} f$, (assuming the combination is a product).
This step can be done in time and space $O(t^{|\psi(u)|})$.
%forrelational specification of functions.
In particular, the tightness of the product function is $O(t^{|\psi(u)|})$.
\item For each neighbor  $c$ of $u$, $c \neq v$ iterate the following 
%(assuming the summation marginalization operator):
%
$
h(u)  \leftarrow h(u)  \cdot  \sum_{\chi(u) \cap
\chi(c)} m_{(c,u)}
$.
%$
%h(u)  \leftarrow h(u)  \bigotimes  \Downarrow_{\chi(u) \cap
%\chi(c)} m_{(c,u)}
%$.
This step can be accomplished in  $O(deg \cdot  hw \cdot \log  t
\cdot t^{hw})$ time and $O(t^{hw})$ space, $deg$ can be dropped if messages are  sent only from leaves to root. %, as we elaborate next.
\item Make $m_{(u,v)} \leftarrow h(u)$.
\end{enumerate}
\vspace{-2mm}

The following bounds of algorithm $CTE$ as a function of the treewidth and hyperwidth  are  restated from \cite{} the literature, 
with a slight simplification to account for one way message-passing.
%\footnote{a specific message can be answered by messages going in one direction only}.

%%%%%%%%%%%%%%%combined theorems%%%%%%%%%%%
\shrink{
\annie{do we need this paragraph, when reading the above paragraph I thought the next thing I was going to read is the bounds, this paragraph disrupts the flow and just explains the history of the citations which i dont think is necessary- cutting it.}
It was shown in \citet{DBLP:journals/ai/GottlobLS00,DBLP:books/cu/p/GottlobGS14}
that a hypertree decomposition of a constraint problem can be
processed in time exponential in the hypertree width.
The work by \citet{DBLP:journals/ai/KaskDLD05}  extended 
this bound %
to any graphical model having sparse
relational specification, 
that is absorbing relative to 0. The following theorem captures the complexity of $CTE$ as a function of both the treewidth and hypertree width.

}

\begin{theorem} [Graph-based complexity of CTE]\label{thm-complexity}
\citep{DBLP:journals/ai/KaskDLD05} 
Given a graphical model \\ ${\cal M} =  {\langle \XX,
{\bf D}, \F  \rangle}$  
and a hypertree-decomposition $\langle T,\chi, \psi \rangle$, let $n$ be the number of variables in $\XX$, 
%$m$ be the number of
%vertices (or clusters) in the tree decomposition, 
$w$ its treewidth, $hw$ its hypertree width and $t$ its tightness,
%$r$ the number of input functions in $F$,
%$deg$ the maximum degree in %$T$,
and $k$ the maximum domain size
of a variable. 
A sum-product query can be computed by $CTE$  within the two following bounds:\\
%\begin{enumerate}
%\item 
1. as a function of the treewidth:
\vspace{-2mm}
\begin{equation}
O(n \cdot k^{w+1}) \label{eq1} 
~~time~and~~O(n k^{w})~~space
\end{equation}
2. as a function of the hyperwidth:
\begin{equation}
 O(n \cdot  hw \cdot \log t
\cdot t^{hw}) \label{eq4}
~~time~and~~O(t^{hw})~~space
\end{equation}
\label{HTelim-complexity1}
\vspace{-10mm}
\end{theorem}

\subsection{Complexity of AND/OR search}

%%%%%%%%%%%%%%%%%%end text from radu%%%%%%%%%%%%%%%%%%%%%

%Search based algorithms were also shown to be bounded as a function of both the treewidth and hypertree width.
%In particular, 
It was shown that  AND/OR search spaces can also reflect problem decomposition captured explicitly in the search space using AND nodes.  When caching of nodes that root identical search
subspace %is utilized 
via what is known as 
context-identical nodes \citep{DBLP:journals/ai/DechterM07} we get a compact %, we get the {\em
%context-minimal} 
AND/OR search space and it allows bounding the AND/OR search graph size by $O(n k^{w+1})$ where $w$ is the
treewidth of the tree-decomposition of the model %that guides the AND/OR search 
\citep{DBLP:journals/ai/DechterM07}.
It was subsequently shown \citep{DBLP:conf/ecai/DechterOM08} that the hypertree width can also be used to bound the AND/OR search graph size.
%when the functions are sparse. 
Consequently,
search algorithms such as depth-first or
best-first search that traverse the AND/OR search graph 
%(potentially using some memory to perform graph search rather than tree search) 
inherit those bounds, which turns out to be
%. Overall the bounds obtained when using AND/OR graph search are 
very similar to those obtained by $CTE$.
%The bound based on treewidth are the same and the bounds based on hypertree width are slightly simplified: 
Any sum-product query can be answered by traversing the AND/OR search graph in 
  1. $O(n \cdot k^{w+1})$ time and $O(k^{w})$ space, and
  2. in $O(n \cdot t^{hw})$ time and $O(t^{hw})$ space.

%}

%It has been shown:

%\shrink{
\begin{proof}
We can build a pseudo-tree that corresponds to the join-tree
underlying the acyclic hypergraph. The context size of each
variable is at most the scope of each function in the acyclic
graphical model. It is clear that only tuples over the context
variables that appear in the relational specification of the
function will be accounted for in the AND/OR search graph (the rest are
associated with "0", so they are inconsistent, or their
probability is "0"), yielding $O(r \cdot t )$ contexts and
therefore, nodes, in the context-minimal AND/OR search graph.
\end{proof}
%}

%\shrink{
\begin{theorem}[Graph-based complexity of search]\citep{DBLP:journals/ai/DechterM07,DBLP:conf/uai/OttenD08}
\label{thm3}
\\ ${\cal M} =  {\langle \XX,
{\bf D}, \F  \rangle}$  
and a hypertree-decomposition $\tau = \langle T,\chi, \psi \rangle$,  having hypertree width $hw$, treewidth $w$,   $t$ is the tightness, $k$ the domain size,
  and $n$ is the number of
  vertices.  Then,
 % \begin{itemize}
      %\item 
    % 1) The AND/OR search graph guided by $\tau$
 % has a size bounded by  $O(n \cdot k^{w+1})$ and $O(n \cdot t^{hw})$,
  %\item 
  %2) 
  any sum-product query can be answered by traversing the AND/OR search graph in 
  1. $O(n \cdot k^{w+1})$ time and $O(k^{w})$ space, and
  2. in $O(n \cdot t^{hw})$ time and $O(t^{hw})$ space.

  \label{thm-4}
  \end{theorem}
 % }

%\jin{This section needs some clarification. Is AND/OR search an alternative method to CTE for solving the same graphical model query? } \rina{yes.} \jin{Perhaps we don't need such a long section. We could briefly discuss CTE and AND/OR algorithms for answering queries before presenting their complexity, perhaps combining Theorems 1, 2, 3 into a sinlge one? }\rina{I agree. it is hard to do it in a clear manner}
%\begin{theorem}
  %Given a graphical model $\M$, and a hypertree-decomposition
  %having hypertree width $hw$, then there exists a tree $\cal T$ such
  %that the context minimal AND/OR search graph based on $\cal T$
  %is  $O(m \cdot t^{hw})$,  when $t$ bounds the
  %function relational specification size and $m$ is the number of
 % vertices in the decomposition.

 % \label{thm-4}
  %\end{theorem}

%\shrink{ 
Thus, if we have a hypertree decomposition and a tree $\cal T$ that
can guide search algorithms such depth-first or
best-first search with, 
%that perform %context-based 
caching  yielding
%along $\cal T$, 
bounds that are exponentially in the hypertree width. In summary:

%From Theorem \ref{thm-4} we can conclude that
\begin{theorem}
Given a graphical model $\M$, and a hypertree-decomposition
having hypertree width $hw$,
the sum-product query can be answered by searching the AND/OR search graph in $O(n \cdot t^{hw})$ time and $O(t^{hw})$ space when $t$ bounds the
  function tightness and $n$ is the number of variables.
 \end{theorem}
%}

\shrink{
Thus both AND/OR graph search and message passing algorithms such as CTE obey the following bounds. ( For message-passing schemes we can show that $m \cdot deg \leq n$ and treat $hw \cdot log t$ as a constant. )

\begin{corollary}
\label{cor1}
Given a Bayesian network, having
$w$ and $hw$ treewidth and hyperwidth respectively,
inference is $O(n \cdot k^{w+1})$ time and $O(k^w)$  space. Also, inference' time is
$O(n \cdot t^{hw})$ and $O(t^{hw})$ space when the functions are sparse having tightness bounded by $t$. 
\end{corollary}

\begin{proof}
    to complete
\end{proof}
}

\usetikzlibrary{bayesnet}

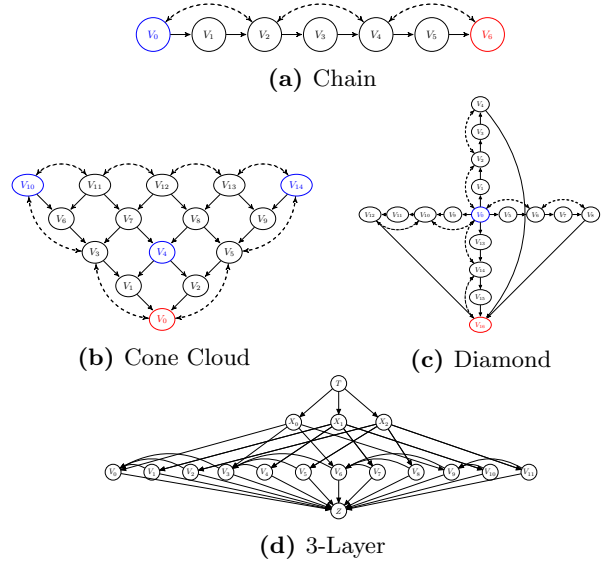
\begin{figure}[t]
     \begin{subfigure}[b]{\columnwidth}
     \centering 
         \resizebox{.6\linewidth}{!}{\begin{tikzpicture}[->,>=stealth',shorten >=1pt,auto,node distance=1.65cm,
  main node/.style={circle,draw,minimum size=1cm,font=\sffamily\small}]
        
            \node[main node, color =blue] (1) { $V_0$};
            \node[main node] (2)[right of =1] { $V_1$};
            \node[main node] (3)[right of =2] { $V_2$};
            \node[main node] (4)[right of=3]{$V_3$};
            \node[main node] (5)[right of=4]{$V_4$};         
            \node[main node] (6)[right of=5]{$V_5$};
            \node[main node, color=red] (7)[right of=6]{$V_6$};

                \draw[thick] (1) edge  (2);
               \draw[thick] (2) edge  (3);
                \draw[thick] (3) edge  (4);
               \draw[thick] (4) edge  (5);
             \draw[thick] (5) edge  (6);
               \draw[thick] (6) edge  (7);
          
              %  \draw[thick] (5) edge  (8);
              %  \draw[thick] (7) edge  (8);
              % \draw[thick] (8) edge  (9);

               \draw[thick, dashed] (1) edge[bend left=45] (3);
            \draw[thick, dashed] (3) edge[bend right=45] (1);

               \draw[thick, dashed] (3) edge[bend left=45] (5);
           \draw[thick, dashed] (5) edge[bend right=45] (3);          
            
            \draw[thick, dashed] (5) edge[bend left=45] (7);
            \draw[thick, dashed] (7) edge[bend right=45] (5);

        \end{tikzpicture}} 
        \caption{Chain}
         \label{fig:chain_7}%
      \end{subfigure}\\
    \begin{subfigure}[b]{.49\columnwidth}
    \centering
         \resizebox{\linewidth}{!}{\begin{tikzpicture}[->,>=stealth',shorten >=1pt,auto,node distance=1.5cm,
  main node/.style={ellipse,minimum size=.5pt ,draw,font=\sffamily\small}]
            \node[main node, color =red] (0) { $V_0$};
            \node[main node] (1)[above left of =0] { $V_1$};
            \node[main node] (2)[above right of =0] { $V_2$};
            \node[main node] (3) [above left of =1] {$V_3$};
            \node[main node,  color =blue] (4) [above right of =1] {$V_4$};
            \node[main node] (5)[above right of =2] { $V_5$};
            \node[main node] (6)[above left of =3] { $V_6$};
            \node[main node] (7) [above right of =3] {$V_7$};
            \node[main node] (8) [above right of =4] {$V_8$};
            \node[main node] (9)[above right of =5] { $V_9$};
            \node[main node,  color =blue] (10)[above left of =6] { $V_{10}$};
            \node[main node] (11) [above right of =6] {$V_{11}$};
            \node[main node] (12) [above right of =7] {$V_{12}$};
           \node[main node] (13)[above right  of=8]{$V_{13}$};
            \node[main node,  color =blue] (14)[above right of =9] { $V_{14}$};

                \draw[thick] (1) edge  (0);
                \draw[thick] (2) edge  (0);
                \draw[thick] (3) edge  (1);
                \draw[thick] (4) edge  (1);
                \draw[thick] (4) edge  (2);
               \draw[thick] (5) edge  (2);
                \draw[thick] (6) edge  (3);
               \draw[thick] (7) edge  (3);
               \draw[thick] (7) edge  (4);
               \draw[thick] (8) edge  (4);

               \draw[thick] (8) edge  (5);
               \draw[thick] (9) edge  (5);
               \draw[thick] (10) edge  (6);
               
               \draw[thick] (11) edge  (6);
               \draw[thick] (11) edge  (7);
               \draw[thick] (12) edge  (7);

                \draw[thick] (12) edge  (8);
               \draw[thick] (13) edge  (8);
                \draw[thick] (13) edge  (9);
               \draw[thick] (14) edge  (9);

               \draw[thick, dashed] (0) edge[bend right=40] (5);
            \draw[thick, dashed] (5) edge[bend left=40] (0);
            
            \draw[thick, dashed] (5) edge[bend right=40] (14);
            \draw[thick, dashed] (14) edge[bend left=40] (5);

            \draw[thick, dashed] (0) edge[bend left=40] (3);
            \draw[thick, dashed] (3)edge[bend right=40] (0);  
            
            \draw[thick, dashed] (3) edge[bend left=40] (10);
            \draw[thick, dashed] (10)edge[bend right=40] (3);

            %%%%%top layer
            \draw[thick, dashed] (11) edge[bend right=50] (10);
            \draw[thick, dashed] (10) edge[bend left=50] (11);  
            
            \draw[thick, dashed] (11) edge[bend left=50] (12);
            \draw[thick, dashed] (12) edge[bend right=50] (11); 
            
            \draw[thick, dashed] (12) edge[bend left=50] (13);
            \draw[thick, dashed] (13) edge[bend right=50] (12);
            
            \draw[thick, dashed] (14) edge[bend right=50] (13);
            \draw[thick, dashed] (13) edge[bend left=50] (14);

        \end{tikzpicture}} 
        \caption{Cone Cloud}
         \label{fig:cone_cloud}%
      \end{subfigure}
     \begin{subfigure}[b]{.49\columnwidth} \centering
        \resizebox{.8\linewidth}{!}{\begin{tikzpicture}[->,>=stealth',shorten >=1pt,auto,node distance=1.25cm,
  main node/.style={ellipse,minimum size=.5pt ,draw,font=\sffamily\small}]
            \node[main node, color =blue] (0) { $V_0$};
            \node[main node] (1)[above of =0] { $V_1$};
            \node[main node] (2)[above of =1] { $V_2$};
            \node[main node] (3) [above of =2] {$V_3$};
            \node[main node] (4) [above of =3] {$V_4$};
            \node[main node] (5)[right of =0] { $V_5$};
            \node[main node] (6)[right of =5] { $V_6$};
            \node[main node] (7) [right of =6] {$V_7$};
            \node[main node] (8) [right of =7] {$V_8$};
            \node[main node] (9)[left of =0] { $V_9$};
            \node[main node] (10)[left of =9] { $V_{10}$};
            \node[main node] (11) [left of =10] {$V_{11}$};
            \node[main node] (12) [left of =11] {$V_{12}$};
           \node[main node] (13)[below of=0]{$V_{13}$};
            \node[main node] (14)[below of =13] { $V_{14}$};
            \node[main node] (15) [below of =14] {$V_{15}$};
           \node[main node, color=red] (16) [below of =15] {$V_{16}$};

                \draw[thick] (0) edge  (1);
                \draw[thick] (0) edge  (5);
                \draw[thick] (0) edge  (9);
                \draw[thick] (0) edge  (13);

                \draw[thick] (1) edge  (2);
               \draw[thick] (2) edge  (3);
                \draw[thick] (3) edge  (4);
               \draw[thick] (5) edge  (6);
               \draw[thick] (6) edge  (7);
               \draw[thick] (7) edge  (8);

               \draw[thick] (9) edge  (10);
               \draw[thick] (10) edge  (11);
               \draw[thick] (11) edge  (12);
               
               \draw[thick] (13) edge  (14);
               \draw[thick] (14) edge  (15);
               \draw[thick] (15) edge  (16);

                \draw[thick] (8) edge  (16);
               \draw[thick] (12) edge  (16);
               \draw[thick] (4) edge [bend left=40]  (16);

               \draw[thick, dashed] (0) edge[bend left=45] (2);
            \draw[thick, dashed] (2) edge[bend right=45] (0);
            
            \draw[thick, dashed] (2) edge[bend left=45] (4);
            \draw[thick, dashed] (4) edge[bend right=45] (2);

            \draw[thick, dashed] (0) edge[bend left=45] (6);
            \draw[thick, dashed] (6)edge[bend right=45] (0);  
            
            \draw[thick, dashed] (6) edge[bend left=45] (8);
            \draw[thick, dashed] (8)edge[bend right=45] (6);        
            
            \draw[thick, dashed] (0) edge[bend left=45] (10);
            \draw[thick, dashed] (10) edge[bend right=45] (0);  
            
            \draw[thick, dashed] (10) edge[bend left=45] (12);
            \draw[thick, dashed] (12) edge[bend right=45] (10); 
            
            \draw[thick, dashed] (0) edge[bend right=45] (14);
            \draw[thick, dashed] (14) edge[bend left=45] (0);
            \draw[thick, dashed] (14) edge[bend right=45] (16);
            \draw[thick, dashed] (16) edge[bend left=45] (14);

        \end{tikzpicture}}  
        \caption{Diamond\\}
           \label{fig:diamond}
      \end{subfigure}
    \shrink{
       \begin{subfigure}[b]{\columnwidth}
    \centering
         \resizebox{.6\linewidth}{!}{\begin{tikzpicture}

% Nodes
\node[latent] (T) {$T$};
\node[latent, below=of T] (X) {$X$};
\node[latent, below=of X, xshift=-8cm] (V0) {$V_0$};
\node[latent, below=of X, xshift=-3cm] (V3) {$V_3$};
\node[latent, below=of X, xshift=2cm] (V6) {$V_6$};
\node[latent, below=of X, xshift=7cm] (V9) {$V_9$};

% Parents of V0
\node[latent, right=of V0] (V1) {$V_1$};
\node[latent, right=of V1] (V2) {$V_2$};

% Parents of V3
\node[latent, right=of V3] (V4) {$V_4$};
\node[latent, right=of V4] (V5) {$V_5$};

% Parents of V6
\node[latent, right=of V6] (V7) {$V_7$};
\node[latent, right=of V7] (V8) {$V_8$};

% Parents of V9
\node[latent, right=of V9] (V10) {$V_{10}$};
\node[latent, right=of V10] (V11) {$V_{11}$};

\node[latent, below=of V5] (Z) {$Z$};

% Edges
\edge {T} {X};  % P(X | T)
% Curved edges
\path[->] (X) edge (V0);  % Curve edge from X to V0
\path[->, bend right] (V1) edge (V0);
\path[->, bend right] (V2) edge (V0);
\path[->, bend right] (V3) edge (V0);

\path[->] (X) edge (V3);  % Curve edge from X to V3
\path[->, bend right] (V4) edge (V3);
\path[->, bend right] (V5) edge (V3);
\path[->, bend right] (V6) edge (V3);

\path[->] (X) edge (V6);  % Curve edge from X to V6
\path[->, bend right] (V7) edge (V6);
\path[->, bend right] (V8) edge (V6);
\path[->, bend right] (V9) edge (V6);

\path[->] (X) edge (V9);  % Curve edge from X to V9
\path[->, bend right] (V10) edge (V9);
\path[->, bend right] (V11) edge (V9);

\edge {V0, V1, V2, V3, V4, V5, V6, V7, V8, V9, V10, V11} {Z};  % P(Z | V0,...,V11)

\end{tikzpicture}} 
        \caption{2-Layer}
         \label{fig:2layer}%
      \end{subfigure}
      }
    \begin{subfigure}[b]{\columnwidth}
    \centering
         \resizebox{.7\linewidth}{!}{\begin{tikzpicture}
% Nodes
\node[latent] (T) {$T$};

\node[latent, below=of T, xshift=-2cm] (X0) {$X_0$};
\node[latent, below=of T, xshift=0cm] (X1) {$X_1$};
\node[latent, below=of T, xshift=2cm] (X2) {$X_2$};

\node[latent, below=of X0, xshift=-8cm, yshift=-0.5cm] (V0) {$V_0$};
\node[latent, below=of X0, xshift=-3cm, yshift=-0.5cm] (V3) {$V_3$};
\node[latent, below=of X0, xshift=2cm, yshift=-0.5cm] (V6) {$V_6$};
\node[latent, below=of X0, xshift=7cm, yshift=-0.5cm] (V9) {$V_9$};

\node[latent, below=of V6] (Z) {$Z$};

% Parents of V0
\node[latent, right=of V0] (V1) {$V_1$};
\node[latent, right=of V1] (V2) {$V_2$};

% Parents of V3
\node[latent, right=of V3] (V4) {$V_4$};
\node[latent, right=of V4] (V5) {$V_5$};

% Parents of V6
\node[latent, right=of V6] (V7) {$V_7$};
\node[latent, right=of V7] (V8) {$V_8$};

% Parents of V9
\node[latent, right=of V9] (V10) {$V_{10}$};
\node[latent, right=of V10] (V11) {$V_{11}$};

% Edges for X_0 dependencies
\edge {T} {X0, X1, X2};  % P(X0 | T), P(X1 | T), P(X2 | T)
\path[->] (X0) edge (V0);  % Curve edge from X to V0
\path[->] (X1) edge (V1);  % Curve edge from X to V0
\path[->] (X2) edge (V2);  % Curve edge from X to V0
\path[->, bend right] (V1) edge (V0);
\path[->, bend right] (V2) edge (V0);
\path[->, bend right] (V3) edge (V0);

\path[->] (X0) edge (V3);  % Curve edge from X to V3
\path[->] (X1) edge (V4);  % Curve edge from X to V0
\path[->] (X2) edge (V5);  % Curve edge from X to V0
\path[->, bend right] (V4) edge (V3);
\path[->, bend right] (V5) edge (V3);
\path[->, bend right] (V6) edge (V3);

\path[->] (X0) edge (V6);  % Curve edge from X to V6
\path[->] (X1) edge (V7);  % Curve edge from X to V0
\path[->] (X2) edge (V8);  % Curve edge from X to V0
\path[->, bend right] (V7) edge (V6);
\path[->, bend right] (V8) edge (V6);
\path[->, bend right] (V9) edge (V6);

\path[->] (X0) edge (V9);  % Curve edge from X to V9
\path[->] (X1) edge (V10);  % Curve edge from X to V0
\path[->] (X2) edge (V11);  % Curve edge from X to V0
\path[->, bend right] (V10) edge (V9);
\path[->, bend right] (V11) edge (V9);

% Edges for X_1 dependencies
\edge {X1} {V1, V4, V7, V10};  % P(V1 | X1), P(V4 | X1), P(V7 | X1), P(V10 | X1)

% Edges for X_2 dependencies
\edge {X2} {V2, V5, V8, V11};  % P(V2 | X2), P(V5 | X2), P(V8 | X2), P(V11 | X2)

% Edge for Z
\edge {V0, V1, V2, V3, V4, V5, V6, V7, V8, V9, V10, V11} {Z};  % P(Z | V0,...,V11)

\end{tikzpicture}} 
        \caption{3-Layer}
         \label{fig:3layer}%
      \end{subfigure}
    \caption{Causal Graphs}
      \label{fig:all_models}
      \vspace{-4mm}
\end{figure}

\section{Causal Effect Estimand Evaluation}
We use these results to analyze 
the plug-in evaluation of causal queries' estimands.
The notion of
empirical Bayesian networks is central to this evaluation.
%associated with a DAG $G$ and whose CPTs are the empirical CPTs generated from a dataset $D$. As a Bayesian network with sparse functions, it obeys complexity bounds based on treewidth and hyperwidth. We will then show that evaluating an estimand can be cast as inference over an empirical Bayesian networks over a DAG that can be extracted from the estimand's expression.
 
\paragraph{Empirical Bayesian Network.} 
Given a directed graph $G$ whose nodes are discrete variables 
 %$BN= \\<X,D,G,\{P_1,…,P_n\}>$  %defined over variables  
 ${\bold X} = \{ X_1,…,X_n\}$ 
 %where $G$ is a directed graph 
 and given a data set $\D = \{d_1,…d_t \} $ over the variables, 
the {\it empirical Bayesian network},
 $empBN(G,\D)$, is the Bayesian network (BN) whose graph is $G$ and its functions are the {\it empirical CPTs} 
extracted from the dataset $\D$.
%over its families (i.e., each variable and its parents in the graph). 
That is, any entry  $(x,pa_X)$ in the empirical CPT  $P_{\D} (X=x|PA_X=pa_X)$ for a variable $X$ and its parents $PA_X$ in the $G$
%distribution over all the variables is determined by counting the frequency of each configuration and normalizing by the data size. Subsequently, any probabilistic quantity over a subset of the variables can be obtained by marginalization. 
is obtained by counting the number of appearances of $(x, pa_X)$ in $\D$, divided by the number of appearances of $pa_X$ in $\D$. 
Formally,
$P_{\D}( x | pa_X) =  \frac{\#\D(x,pa_X)}{\#\D(pa_X)}$
where $\#\D(s)$ is the number of elements in $\D$ that are consistent with $s$.

%divided by the frequencies and normalizing by the data size. Subsequently, any probabilistic quantity over a subset of the variables can be obtained by marginalization.
%\end{definition}

%counting the frequency of each confThus, for each variable and its parents we can compute the empirical conditional probability tables and the product of these tables yields the empirical distribution over all the variables. \rina{there is subtleties here: that empirical distribution obtained from the data, vs the one obtained from the empirical Bayesian network are not the same... we need to revise}

It is easy to see that the number of non-zero configurations of any CPT table of $empBN(G,\D)$
cannot exceed the data size $t$.
 Therefore, by Theorem~\ref{thm-complexity}, we can immediately conclude the following.
% \annie{t is first defined as tightness and now its the data size?}\rina{If data size is t than tightness is bounded by t.}
 %from Theorem 
%\ref{HTelim-complexity1} and from the discussion on AND/OR search that

%\shrink{
\begin{theorem}[Complexity of empirical BN.]
\label{thm3a} Assume a given  DAG $G$ over variables $\{X_1,...,X_n\}$ and let $hw$ and $w$ be the hypertree width and treewidth of the Bayesian network,
%hypergraph $<G,S>$,
respectively. 
Given also
a dataset $\D$, where $t = |\D|$, and domain sizes bounded by $k$, then computing a sum-product query over $empBN(G,\D)$ by $CTE$ 
%or by AND/OR search 
as a function of $w$ 
is $O(n  \cdot k^{w+1})$ time and
$O(k^{w})$ space. Computing the query as a function of $hw$  is 
$O(n \cdot  hw \cdot \log t
\cdot t^{hw})$, time and
 $ O(t^{hw})$ space by $CTE$.  
  %it is 
%$O(n \cdot t^{hw})$, time and  
% $O(t^{hw})$, space by AND/OR search.
%\end{itemize}
\end{theorem}

\shrink{
Computing 
%\jin{needs to specify "inference".} 
\jin{need revision}
\begin{itemize}
    \item 
$O(n  \cdot k^{w+1})$ time and space 
$O(k^{w})$,
\item $O(n \cdot  hw \cdot \log t
\cdot t^{hw})$, and
 $ O(n \cdot  t^{hw})$ 
 \item and by search it is 
$O(n \cdot t^{hw})$, time and  
 $O(t^{hw})$, space.
\end{itemize}
}

\shrink{
\begin{proof}
    $empBN(G,D)$ is a Bayesian network having tightness $t$, and for Baesian network $r=n$, and $m$ is bounded by $n$, so the claim follows immediately from Theorems 
 \ref{thm-cte-complexity} and \ref{HTelim-complexity1}.

%\annie{ need clarification here, I dont think its clear or how its simplified to $n$ } \rina{For Bayesian networks r, the number of functions,  equals n, the number of variables.} \annie{ yes O(r+N) = O(2n) =O(n) but for the hw bound you have m · deg · hw · log t, not clear how that simplifies to n }
\end{proof}
  }

\begin{corollary} If $empBN$ has hypertree width 1, the sum-product query  can be answered in almost linear time, $O(n \cdot t\log t)$.
\end{corollary}
%\jin{Acyclic means polytree?}

%\jin{Do we really need the concept of (empirical) Bayesian network? An estimand specifies a set of variables and functions, and a query, which naturally maps to the concept of a graphical model. No need to introduce Bayesian network as an intermediate concept. We could directly define an empirical graphical model that corresponds to an estimand instead. } \rina{I think it is possible. I think the notion on of emprBN is of interest.}

\paragraph{Implication to estimands evaluation.} 
%An estimand expression is typically a product of sums of products. It can also be a ratio of such expressions. 
%We next associate estimands with graphs.
%\paragraph{Estimands hypergraphs.} 
We show a causal effect estimand can be associated with a hierarchy of empirical Bayesian networks so that evaluating the sum-product inference on each will yield the estimand evaluation. Applying a hypertree decomposition algorithm like $CTE$ to each sub-expression will yield an effective estimand evaluation scheme whose performance would be bounded by the  graph parameters of treewidth and hyperwidth.
We  first illustrate via several examples.

\begin{figure}[t]
\begin{subfigure}[t]{\columnwidth}
    \resizebox{\linewidth}{!}{\usetikzlibrary{positioning}
\begin{tikzpicture}[-,>=stealth',shorten >=1pt,auto,node distance=2.5cm,
  main node/.style={ellipse,draw,font=\sffamily\small}]
        
            \node[main node] (1) at (3,-1.5) {$V'_0, V_1, V_2, V_3, V_4, V_5, V_6$};
           \node[above=0.1cm of 1,xshift=1cm, blue] {$F(V'_0, V_1, V_2, V_3, V_4, V_5)$}; 
            \node[main node] (2) at (0, 0)  {$V_0, V_1, V_2, V_3, V_4, V_5$};
            \node[above=0.1cm of 2,xshift=0cm, blue] {$F(V_0, V_1, V_2, V_3, V_4, V_5)$}; 
            \node[main node] (3)at (0,-3)  {$V'_0, V_1, V_2, V_3, V_4$};
            \node[above=0cm of 3,xshift=-.5cm, blue] {$F(V'_0, V_1, V_2, V_3, V_4)$}; 
            \node[main node] (4) at (5,-3) {$V'_0, V_1, V_2$};
            \node[above=0.1cm of 4,xshift=0cm, blue] {$F(V'_0, V_1, V_2)$}; 
            
            \node[main node] (5) at (7,-2) {$V'_0$};  
            \node[above=0.1cm of 5,xshift=0cm, blue] {$F(V'_0)$}; 
           
            \node[main node] (6) at (-2.5, -1) {$V_0, V_1, V_2, V_3$};   
             \node[above=0.1cm of 6,xshift=-1cm, blue] {$F(V_0, V_1, V_2, V_3)$};
            
            \node[main node] (7)at (-.25, -1.5) {$V_0, V_1$};
             \node[above=0.0cm of 7,xshift=.8cm, blue] {$F(V_0, V_1)$};            
                \draw[thick] (1) -- (2);
 
                \draw[thick] (1) -- (4); 
                
                \draw[thick] (1) -- (3)  ; 
                
                \draw[thick] (4) -- (5);   

                \draw[thick] (2) -- (6) ; 
     
                \draw[thick] (2) -- (7) ;

        \end{tikzpicture}} 
\end{subfigure}
\caption{Hypertree for Eqn
\ref{eq:smallchainexp_flattened} with $hw=1$ %and largest cluster $F_4$ containing all the variables.
}
\vspace{-2mm}
    \label{fig:chain_HT}%
\end{figure}
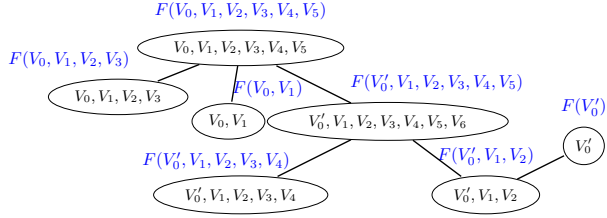
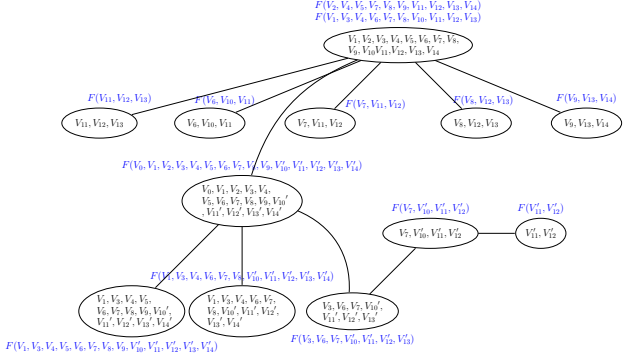
\begin{figure}[t]
	\centering
    \resizebox{\linewidth}{!}{\begin{tikzpicture}[-,>=stealth',shorten >=1pt,auto,node distance=5.5cm,
  main node/.style={ellipse,draw,minimum size=1.5cm,font=\sffamily\Large}]
        
            \node[main node,text width=5cm] (1) { $V_1,V_2,V_3, V_4, V_5, V_6, V_7,V_8,$\\$ V_9, V_{10} V_{11},$$V_{12}, V_{13}, V_{14}$};  
            \node[above=0.1cm of 1,xshift=0cm, blue] {\shortstack{\Large$F(V_2, V_4, V_5, V_7, V_8, V_9, V_{11}, V_{12}, V_{13}, V_{14})$\\\Large$F(V_1, V_3, V_4, V_6, V_7, V_8, V_{10}, V_{11}, V_{12}, V_{13})$}}; 
            
            \node[main node] (3)[ below right of =1] { $V_8, V_{12}, V_{13}$};      
            \node[above=0.0cm of 3,xshift=0.35cm, blue] {\Large$F(V_8, V_{12}, V_{13})$}; 
            
            \node[main node] (4)[right of=3]{$V_9, V_{13}, V_{14}$};
            
            \node[above=0.1cm of 4,xshift=0cm, blue] {\Large$F(V_9, V_{13}, V_{14})$};

            \node[main node] (5)[below left of=1]{$V_7, V_{11}, V_{12}$};    
            \node[above=-.2cm of 5,xshift=2.75cm, blue] {\Large$F(V_7, V_{11}, V_{12})$};

            \node[main node] (6)[left of=5]{$V_6, V_{10}, V_{11}$};
            \node[above=0cm of 6,xshift=0.75cm, blue] {\Large$F(V_6, V_{10}, V_{11})$};

            \node[main node] (7)[left of=6]{$V_{11}, V_{12}, V_{13}$};
            \node[above=0.1cm of 7,xshift=1cm, blue] {\Large$F(V_{11}, V_{12}, V_{13})$}; 
            
            \node[main node] (8)[below left of =5,text width=4cm] { $V_0,V_1, V_2, V_3, V_4,$\\$ V_5, V_6, V_7, V_8, V_9, {{V_{10}}’}$\\$, {{V_{11}}’},{{V_{12}}’}, {{V_{13}}’}, {{V_{14}}’}$};
            \node[above=0.1cm of 8,xshift=0cm, blue] {\Large$F(V_0, V_1, V_2, V_3, V_4, V_5, V_6, V_7, V_8, V_9, V_{10}', V'_{11}, V'_{12}, V'_{13}, V_{14}')$};

 \node[main node] (10)[below of =8, text width=3.5cm] { $V_1,  V_3, V_4, V_6, V_7,$\\$ V_8, {{V_{10}}’}, {{V_{11}}’}, {{V_{12}}’},$\\$ {{V_{13}}’}, {{V_{14}}’}$};  
\node[above=0.1cm of 10,xshift=0cm, blue] {\Large$F( V_1, V_3, V_4, V_6, V_7, V_8, V_{10}', V'_{11}, V'_{12}, V'_{13}, V_{14}')$}; 

\node[main node] (9)[ left of =10, text width=3.5cm] { $V_1, V_3, V_4, V_5,$\\$ V_6, V_7, V_8, V_9,{{V_{10}}’},$\\$ {{V_{11}}’}, {{V_{12}}’}, {{V_{13}}’}, {{V_{14}}’}$};
\node[below=0.1cm of 9,xshift=-1cm, blue] {\Large$F( V_1, V_3, V_4, V_5, V_6, V_7, V_8, V_9, V_{10}', V'_{11}, V'_{12}, V'_{13}, V_{14}')$}; 
             
  \node[main node] (11)[ right of=10, text width=3cm]{$V_3, V_6, V_7, {{V_{10}}’}, $\\${{V_{11}}’}, {{V_{12}}’}, {{V_{13}}’}$};
\node[below=0.1cm of 11,xshift=0cm, blue] {\Large$F( V_3, V_6, V_7, V_{10}', V'_{11}, V'_{12}, V'_{13})$};

  \node[main node] (12)[above right of=11]{$V_{7}, V'_{10}, V'_{11}, V'_{12}$};
  \node[above=0.1cm of 12,xshift=0cm, blue] {\Large$F( V_{7}, V'_{10}, V'_{11}, V'_{12})$}; 
  
    \node[main node] (13)[ right of=12]{$V'_{11}, V'_{12}$};
    \node[above=0.1cm of 13,xshift=0cm, blue] {\Large$F(V'_{11}, V'_{12})$};

               % \draw[thick] (1) edge  (2);
               \draw[thick] (1) edge  (3);
                \draw[thick] (1) edge  (4);
              % \draw[thick] (1) edge[bend right=45]  (8);

             \draw[thick] (1) edge  (5);
               \draw[thick] (1) edge  (6);
                    \draw[thick] (1) edge  (7);
                    \draw[thick] (1) edge[bend right=25]  (8);    

                  \draw[thick] (8) edge  (9);
                    \draw[thick] (8) edge  (10);
                    \draw[thick] (8) edge[bend left=35]  (11);         \draw[thick] (11) edge  (12);
                    \draw[thick] (12) edge  (13);
                  
        \end{tikzpicture}}  
	\caption{Cone-cloud Estimand's Hypertree with $hw=2$
 }
	\label{fig:cone_ht}
 \vspace{-2mm}
\end{figure}

\begin{example}\label{ex1}
Consider the model in Figure \ref{fig:chain_7}. To evaluate the query $P(V_6 \mid do(V_0))$, the ID algorithm \citep{jinthesis,shpitser2006identification} generates the expression:
%\vspace{mm}
%\jin{why use lower case $v_0$?}% 
\begin{small}
\begin{multline} \label{eq:smallchainexp}
    P(V_6 \mid do(V_0))
    = \!\!\!\! \sum\limits_{V_1, V_2, V_3, V_4, V_5} \!\!\!\!\!\! P(V_5 | V_0, V_1, V_2, V_3, V_4) \times \\
    P (V_3 | V_0, V_1, V_2) \, P(V_1 | V_0) 
    \sum\limits_{V_0} P(V_6 | V_0, V_1, V_2, V_3, V_4, V_5 ) \\  
    \times P(V_4 | V_0, V_1, V_2, V_3) P(V_2 | V_0, V_1) \,P(V_0). 
\end{multline}
%\vspace{-1mm}
\end{small}
By variable renaming of the summation variables in the second sum  and moving all summations to the head of the expression we get an equivalent expression (in this case we rename $v_0$  to $v'_0$), yielding: \vspace{-5mm}

%\annie{ can we condense the two equations to one step, "by variable renaming and moving all summations to the front"} \rina{yes.}
%\begin{small}
%\begin{multline} \label{eq:smallchainexp1}
%    P(V_6 \mid do(V_0))
 %   = \!\!\!\! \sum\limits_{V_1, V_2, V_3, V_4, V_5} \!\!\!\!\!\! P(V_5 | V_0, V_1, V_2, V_3, V_4) \times \\
%     P (V_3 | V_0, V_1, V_2) \, P(V_1 | V_0) 
%     \sum\limits_{V'_0} P(V_6 | V'_0, V_1, V_2, V_3, V_4, V_5 ) \\ \times P(V_4 | V'_0, V_1, V_2, V_3) P(V_2 | V'_0, V_1) \,P(V'_0).   
%\end{multline}
%\end{small}

%\vspace{.4in}
%By moving all summations to the head of the %expression we get the flat sum product:
\begin{small}
\begin{multline} \label{eq:smallchainexp_flattened}
    P(V_6 \mid do(V_0))
    = \!\!\!\! \sum\limits_{V_1, V_2, V_3, V_4, V_5, v'_0} \!\!\!\!\!\! P(V_5 | V_0, V_1, V_2, V_3, V_4) \\
    \times P (V_3 | V_0, V_1, V_2) \, P(V_1 | V_0)P(V_6 | V'_0, V_1, V_2, V_3, V_4, V_5 ) \\ \times   P(V_4 | V'_0, V_1, V_2, V_3) \,P(V_2 | V'_0, V_1) \,P(V'_0).  
\end{multline}
\end{small}
\vspace{-6mm}

The latter expression corresponds to  a sum-product reasoning task over a Bayesian network defined by the 7 CPTs in Expression (\ref{eq:smallchainexp_flattened}). 
A hypertree decomposition with $hw=1$ is given in Figure \ref{fig:chain_HT}.
\shrink{
Treating the scopes of the 7 CPTs as nodes in a dual graph, it is easy to see that they have an underlying hypertree-decomposition 
%( a {\it join-tree}) which means that the hypergraph of the Nayesian network is actually a hypertree, 
as shown in Figure \ref{fig:chain_HT1}. 
}
%this is an acyclic BN having $hw=1$, 
By Theorem \ref{thm3a}, 
the expression can be computed by $CTE$ in time
$O(n \cdot  log t \cdot t)$ 
%and by AND/OR search by $O(n  \cdot t)$  
where $t$ is the data size (so $t$ bounds the tightness of the Empirical BN). 
In contrast the treewidth of this tree-decomposition has $w=6$, a potentially loose bound. Generally, for chains of length $n$, the treewidth  is $w=n-1$, leading to an exponential bound of $O(n \cdot k^n)$ while the hyperwidth stays $hw=1$. Clearly, the hyperwidth provides a far tighter and more informative bound for large chains.
\end{example}
\vspace{-2mm}

%Our next example involves a more complex expression which possess a hyperwidth of size 2.
\begin{example} 
Consider the estimand expression in Eq.~\eqref{eq:cone}, obtained from a 15 variable cone-cloud graph (Figure~\ref{fig:cone_cloud}). 
%\jin{Why are $V_{10}', V_{14}'$ in Eq.~(\ref{eq:cone})  already renamed but others not?} \rina{Annie, explained that this is how the expression looks like. For variables appearing in the do(x,y,z) they distinguish the constant assigned in the intervened variable and when those are free variables in the expression.} It is composed of the product of 2 sums.
%\jin{do you mean "sums"?}, 
%The first sum from the left is called the ``primary" sum. % and any other sum is termed %``inner". 
Once we rename the variables in the inner sums and move the summation variables to the head of the first sum, we get the expression Eq.~\eqref{eq:cone_flattened}.
Here  Eq.~\eqref{eq:cone_flattened} can be viewed as a sum-product inference over a probabilistic graphical model.
 %  A dual graph
  %  is depicted in Figure \ref{fig:cone_dual}.
   % \jin{?. This is not a dual graph.}\rina{it is a join-graph. }. 
    In this case however the hypergraph is not a hypertree;
    %it has a cycle and no edge can be eliminated while maintaining the connectedness property. 
    but it can be embedded in a hypertree 
    %by combining nodes 1 and 2 
   % it has a hypertree decomposition -decomposition 
    having $hw=2$, depicted in Figure \ref{fig:cone_ht} (the top cluster includes 2 functions).
Therefore the complexity of evaluating this expression by $CTE$ is  $O(n \cdot \log t \cdot t^2)$. 
%or  $O(n \cdot  t^2)$ by AND/OR search.  
The treewidth is $w=14$ (there is a cluster with 15 variables) leading to a bound of $O(n k^{15})$ time.  
%\annie{how did you get 10 ? isnt it 14 ?, theres a %cluster with 15 variables$ V_0..V_{14}$}
\vspace{-10mm}
%%%%%%%%%%%%%%%%%%%%%%%%
\begin{strip}
\begin{small}
    \begin{multline}\label{eq:cone}
  P(V_0 |do( V_{14}, V_{10}, V_4 )) = 
    \sum_{V_1, V_2, V_3, V_5, V_6, V_7, V_8, V_9, V_{11}, V_{12}, V_{13}} 
    P ( V_2 |  V_4, V_5, V_7, V_8, V_9, V_{11}, V_{12}, V_{13}, V_{14}) P(V_9|V_{13}, V_{14}) \times \\
      P(V_8|V_{12},V_{13})P(V_1| V_3, V_4, V_6, V_7, V_8, V_{10}, V_{11}, V_{12}, V_{13} )
    P(V_7| V_{11}, V_{12})  P(V_6 | V_{10},V_{11} ) P(V_{11}, V_{12}, V_{13})\times\\
    \sum\limits_{V_{10}', V_{11}, V_{12}, V_{13}, V_{14}'} P(V_0 | V_1, V_2, V_3, V_4, V_5, V_6, V_7, V_8, V_9, V_{10}', V_{11}, V_{12}, V_{13}, V_{14}') P(V_3, V_{13} | V_{6}, V_{7}, V_{10}', V_{11}, V_{12})P(V_{11}, V_{12}) \times \\
    P(V_5 | V_1, V_3, V_4, V_6, V_7, V_8, V_9, V_{10}', V_{11}, V_{12}, V_{13}, V_{14}')   P(V_{14}' | V_1, V_3, V_4, V_6, V_7, V_8, V_{10}', V_{11},V_{12}, V_{13}) P(V_{10}' | V_{7}, V_{11}, V_{12}) 
\end{multline}
\end{small}
\vspace{-17mm}
\end{strip}% Switch back to two-column mode
\begin{strip}
\begin{small}
\begin{multline}\label{eq:cone_flattened}
    P(V_0 | do( V_{14}, V_{10}, V_4) )= 
    \sum\limits_{V_1, V_2, V_3, V_5, V_6, V_7, V_8, V_9, V_{11}, V_{12}, V_{13},V'_{10},V'_{11},V'_{12},V'_{13},V'_{14}} P ( V_2 |  V_4, V_5, V_7, V_8, V_9, V_{11}, V_{12}, V_{13}, V_{14}) \\
    \times P(V_9|V_{13}, V_{14}) P(V_8|V_{12},V_{13}) P(V_1| V_3, V_4, V_6, V_7, V_8, V_{10}, V_{11}, V_{12}, V_{13} )
    P(V_7| V_{11}, V_{12}) P(V_6 | V_{10},V_{11} )  P(V_{11}, V_{12}, V_{13}) \\
    \times P(V_0 | V_1, V_2, V_3, V_4, V_5, V_6, V_7, V_8, V_9, V_{10}', V'_{11}, V'_{12}, V'_{13}, V_{14}') P(V_5 | V_1, V_3, V_4, V_6, V_7, V_8, V_9, V_{10}', V'_{11}, V'_{12}, V'_{13}, V_{14}') \\ \times P(V_3, V'_{13} | V_{6}, V_{7}, V_{10}', V'_{11}, V'_{12})  P(V_{14}' | V_1, V_3, V_4, V_6, V_7, V_8, V_{10}', V'_{11},V'_{12}, V'_{13}) 
     P(V'_{10} | V_{7}, V'_{11}, V'_{12}) P(V'_{11}, V'_{12}) 
\end{multline}
\end{small}
\vspace{-5mm}
\end{strip}% Switch back to two-column mode

%%%%%%%%%%%%%%%%%%%%%%%%%%%%
\shrink{
\begin{multline}\label{eq:cone_flattened}
    P(V_0 |do( V_{14}, V_{10}, V_4 )) = 
    \sum\limits_{V_1, V_2, V_3, V_5, V_6, V_7, V_8, V_9, V_{11}, V_{12}, V_{13}} P ( V_2 |  V_4, V_5, V_7, V_8, V_9, V_{11}, V_{12}, V_{13}, V_{14})
    P(V_9|V_{13}, V_{14}) \\ \times  P(V_8|V_{12},V_{13})P(V_1| V_3, V_4, V_6, V_7, V_8, V_{10}, V_{11}, V_{12}, V_{13} )
    P(V_7| V_{11}, V_{12}) \\ \times P(V_6 | V_{10},V_{11} ) P(V_{11}, V_{12}, V_{13}) \times\\
    \sum\limits_{V_{10}', V_{11}, V_{12}, V_{13}, V_{14}'} P(V_0 | V_1, V_2, V_3, V_4, V_5, V_6, V_7, V_8, V_9, V_{10}', V_{11}, V_{12}, V_{13}, V_{14}') \times \\
    P(V_5 | V_1, V_3, V_4, V_6, V_7, V_8, V_9, V_{10}', V_{11}, V_{12}, V_{13}, V_{14}')P(V_3, V_{13} | V_{6}, V_{7}, V_{10}', V_{11}, V_{12})  \\ \times  P(V_{14}' | V_1, V_3, V_4, V_6, V_7, V_8, V_{10}', V_{11},V_{12}, V_{13}) P(V_{10}' | V_{7}, V_{11}, V_{12}) P(V_{11}, V_{12}) \hspace{10pt}
\end{multline}
}
\shrink{ 
\begin{multline}\label{eq:cone1}
    P(V_0 | do(V_{14}, V_{10}, V_4 ) ) = \\
    \sum\limits_{V_1, V_2, V_3, V_5, V_6, V_7, V_8, V_9, V_{11}, V_{12}, V_{13}} P ( V_2 |  V_4, V_5, V_7, V_8, V_9, V_{11}, V_{12}, V_{13}, V_{14})\times \\
    P(V_9|V_{13}, V_{14}) P(V_8|V_{12},V_{13}) P(V_1| V_3, V_4, V_6, V_7, V_8, V_{10}, V_{11}, V_{12}, V_{13} )\times \\
    P(V_7| V_{11}, V_{12}) P(V_6 | V_{10},V_{11} ) P(V_{11}, V_{12}, V_{13}) \times \\
    \sum\limits_{V_{10}', V_{11}, V_{12}, V_{13}, V_{14}'} P(V_0 | V_1, V_2, V_3, V_4, V_5, V_6, V_7, V_8, V_9, V_{10}', V_{11}, V_{12}, V_{13}, V_{14}') \times \\
    P(V_5 | V_1, V_3, V_4, V_6, V_7, V_8, V_9, V_{10}', V_{11}, V_{12}, V_{13}, V_{14}') \times \\P(V_{14}' | V_1, V_3, V_4, V_6, V_7, V_8, V_{10}', V_{11},V_{12}, V_{13}) \times\\
    P(V_3, V_{13} | V_{6}, V_{7}, V_{10}', V_{11}, V_{12})  P(V_{10}' | V_{7}, V_{11}, V_{12}) P(V_{11}, V_{12}) \hspace{10pt}
\end{multline}
}

\shrink{
\begin{strip}
\begin{small}
\begin{multline}\label{eq:2layer}
P(Z | do(T)) = \sum_{V0,V1,V2,V3,V4,V5,V6,V7,V8,V9,V10,V11}P(Z|V0,V1,V2,V3,V4,V5,V6,V7,V8,V9,V10,V11)P(T)\\\sum_{X}P(X|T)P(V0|V1,V2,V3,X)P(V3|V4,V5,V6,X)P(V6|V7,V8,V9,X)P(V9|V10,V11,X)
\end{multline}
\end{small}
\end{strip}%
\begin{strip}
\begin{small}
\begin{multline}\label{eq:3layer_1}
P(Z \mid do(T)) = \sum\limits_{V_0, V_1, V_2, V_3, V_4, V_5, V_6, V_7, V_8, V_9, V_{10}, V_{11}} P(Z \mid V_0, V_1, V_2, V_3, V_4, V_5, V_6, V_7, V_8, V_9, V_{10}, V_{11}) P(T)\\
\sum\limits_{X_0, X_1, X_2} P(X_0 \mid T) P(X_1 \mid T) P(X_2 \mid T)P(V_0 \mid V_1, V_2, V_3, X_0) P(V_3 \mid V_4, V_5, V_6, X_0) P(V_6 \mid V_7, V_8, V_9, X_0) P(V_9 \mid V_{10}, V_{11}, X_0) \\P(V_1 \mid X_1) P(V_4 \mid X_1) P(V_7 \mid X_1) P(V_{10} \mid X_1) P(V_2 \mid X_2) P(V_5 \mid X_2) P(V_8 \mid X_2) P(V_{11} \mid X_2) 
\end{multline}
\end{small}
\end{strip}% 
\begin{strip}
\begin{small}
\begin{multline}\label{eq:3layer_2}
P(\textbf{V} \mid do(T,Z)) = \sum_{X_0, X_1, X_2} P(X_0 \mid T) P(X_1 \mid T) P(X_2 \mid T) P(T) 
 P(V_0 \mid V_1, V_2, V_3, X_0) P(V_3 \mid V_4, V_5, V_6, X_0)\\ P(V_6 \mid V_7, V_8, V_9, X_0) P(V_9 \mid V_{10}, V_{11}, X_0) P(V_1 \mid X_1)  P(V_4 \mid X_1) P(V_7 \mid X_1) P(V_{10} \mid X_1) P(V_2 \mid X_2) P(V_5 \mid X_2) \\
 P(V_8 \mid X_2) P(V_{11} \mid X_2) \hspace{50mm}
\end{multline}
\end{small}
\end{strip}% 

\begin{strip}
\begin{small}
\begin{multline}\label{eq:diamond}
P\left(V_{16} \middle| do(V_0), do(V_4)\right) = \sum_{V_5,V_6,V_7,V_8,V_9,V_{10},V_{11},V_{12},V_{13},V_{14},V_{15}}P\left(V_{15} \middle| V_0,V_{13},V_{14}\right) \times 
\frac{numerator}{denominator}
P\left(V_{13} \middle| V_0\right)P\left(V_{11} \middle| V_0,V_9,V_{10}\right)\\
\times P\left(V_9 \middle| V_0\right)P\left(V_7 \middle| V_0,V_5,V_6\right)P\left(V_5 \middle| V_0\right)
\\
\sum_{V_0'}P\left(V_{12} \middle| V_0',V_9,V_{10},V_{11}\right)
P\left(V_{10} \middle| V_0',V_9\right)P\left(V_0'\right)
\\
\sum_{V_0'}P\left(V_8 \middle| V_0',V_5,V_6,V_7\right)P\left(V_6 \middle| V_0',V_5\right)P\left(V_0'\right)
\end{multline}
\end{small}
\end{strip}% 
\begin{strip}
\begin{small}
\begin{multline}\label{eq:numerator}
numerator= \sum_{V_0',V_2,V_6,V_{10}}P\left(V_{16} \middle| V_0',V_1,V_2,V_3,V_4,V_5,V_6,V_7,V_8,V_9,V_{10},V_{11},V_{12},V_{13},V_{14},V_{15}\right)P\left(V_4,V_{14} \middle| V_0',V_1,V_2,V_3,V_5,V_6,V_7,V_8,V_9,V_{10},V_{11},V_{12},V_{13}\right)\\
P\left(V_2 \middle|V_0',V_1,V_5,V_6,V_7,V_8,V_9,V_{10},V_{11},V_{12},V_{13}\right)
P\left(V_8 \middle| V_0',V_5,V_6,V_7,V_9,V_{10},V_{11},V_{12},V_{13}\right)
P\left(V_6 \middle|V_0',V_5,V_9,V_{10},V_{11},V_{12},V_{13}\right)\\
P\left(V_{12} \middle| V_0',V_9,V_{10},V_{11},V_{13}\right)P\left(V_{10} \middle| V_0',V_9,V_{13}\right)P\left(V_0'\right)\\
\end{multline}
\end{small}
\end{strip}% 
\begin{strip}
\begin{small}
\begin{multline}\label{eq:denominator}
denominator=\sum_{V_0',V_2,V_6,V_{10}}{P\left(V_4 \middle| V_0',V_1,V_2,V_3,V_5,V_6,V_7,V_8,V_9,V_{10},V_{11},V_{12},V_{13}\right)P\left(V_2 \middle| V_0',V_1,V_5,V_6,V_7,V_8,V_9,V_{10},V_{11},V_{12},V_{13}\right)P\left(V_8 \middle| V_0',V_5,V_6,V_7,V_9,V_{10},V_{11},V_{12},V_{13}\right)P\left(V_6 \middle| V_0',V_5,V_9,V_{10},V_{11},V_{12},V_{13}\right)P\left(V_{12} \middle| V_0',V_9,V_{10},V_{11},V_{13}\right)P\left(V_{10} \middle| V_0',V_9,V_{13}\right)P\left(V_0'\right)}
\end{multline}
\end{small}
\end{strip}% 
}

%More generally, for $n$-dimentional 

%cones the treewidth grows linearly with $n$ leading to a bound of $O(n k^{n})$. The $hw$ may increase as well but we speculate that this will be at a far slower pace. \rina{can we make a clearer statement?}
\shrink{
\begin{figure*}[!b]
	\centering
    \resizebox{.7\linewidth}{!}{\begin{tikzpicture}[-,>=stealth',shorten >=1pt,auto,node distance=6cm,
  main node/.style={ellipse,draw,minimum size=1.5cm,font=\sffamily\Large},
    every label/.style={font=\sffamily\Large}]
        
            \node[main node,text width=4cm, label=above:{$ F_1$}] (1) { $V_2, V_4, V_5,V_7,$$V_8,$\\$ V_9, V_{11},$$V_{12}, V_{13}, V_{14}$};  
            \node[main node,text width=3.5cm, label=above left:{$\large F_2$}] (2)[below left of =1] { $V_1, V_3, V_4, V_6, V_7,$\\$ V_8, V_{10}, V_{11}, V_{12}, V_{13}$};
            \node[main node, label=above right:{$F_3$}] (3)[right of =2] { $V_8, V_{12}, V_{13}$};
            \node[main node, label=above right:{$F_4$}] (4)[right of=3]{$V_9, V_{13}, V_{14}$};
            
            \node[main node,label=below right:{$F_5$}] (5)[below left of=2]{$V_7, V_{11}, V_{12}$};
            \node[main node, label=below right:{$F_6$}] (6)[right of=5]{$V_6, V_{10}, V_{11}$};
            \node[main node, label=below right:{$F_7$}] (7)[right of=6]{$V_{11}, V_{12}, V_{13}$};

            \node[main node, label=left:{$F_8$}] (8)[below left of =5,text width=3.5cm] { $V_0,V_1, V_2, V_3, V_4,$\\$ V_5, V_6, V_7, V_8, V_9, {{V_{10}}’}$\\$, {{V_{11}}’},{{V_{12}}’}, {{V_{13}}’}, {{V_{14}}’}$};
            \node[main node, label=below left:{$F_9$}] (9)[below left of =8, text width=3.5cm] { $V_1, V_3, V_4, V_5,$\\$ V_6, V_7, V_8, V_9,{{V_{10}}’},$\\$ {{V_{11}}’}, {{V_{12}}’}, {{V_{13}}’}, {{V_{14}}’}$};
            \node[main node, label=below right:{$F_{10}$}] (10)[right of =9, text width=4cm] { $V_1,  V_3, V_4, V_6, V_7, V_8,$\\$ {{V_{10}}’}, {{V_{11}}’}, {{V_{12}}’},{{V_{13}}’}, {{V_{14}}’}$};          
  \node[main node, label=below right:{$F_{11}$}] (11)[ right of=10, text width=3cm]{$V_3, V_6, V_7, {{V_{10}}’}, $\\${{V_{11}}’}, {{V_{12}}’}, {{V_{13}}’}$};
  \node[main node, label=below right:{$F_{12}$}] (12)[above right of=11]{$V_{7}, V'_{10}, V'_{11}, V'_{12}$};
    \node[main node, label=below right:{$F_{13}$}] (13)[ right of=12]{$V'_{11}, V'_{12},$};

                \draw[thick] (1) -- (2) node[midway, color=blue,fill=white, swap] {\small$V_4, V_7, V_8, V_{11}, V_{12}, V_{13}$};
                \draw[thick] (1) -- (3) node[midway, color=blue,fill=white, swap] {\small$V_8, V_{12}, V_{13}$};
                \draw[thick] (1) -- (4) node[midway, color=blue,fill=white, swap] {\small$V_9, V_{13}, V_{14}$};
\draw[thick, bend right=45] (1) to node[midway, color=blue, fill=white, swap] {\small$V_2, V_4, V_5, V_7, V_8$} (8);
                
                \draw[thick] (2) -- (5) node[midway, color=blue,fill=white, swap] {\small$V_7, V_{11}, V_{12}$};
                \draw[thick] (2) -- (6) node[midway, color=blue,fill=white, swap] {\small$V_6, V_{10}, V_{11}$};
                \draw[thick] (2) -- (7) node[midway, color=blue,fill=white, swap] {\small$V_{11}, V_{12}, V_{13}$};
    
\draw[thick, bend right=25] (2) to node [midway, color=blue, fill=white, swap] {\small$V_1, V_3, V_4, V_6, V_7, V_8$} (8);
                
 \draw[thick] (8) -- (9) node[midway, color=blue, fill=white, swap] 
        {\begin{minipage}{3cm}\centering \small $V_1, V_2, V_3, V_4, V_5,$\\$ V_6, V_7, V_8, V_9,$\\$ {{V_{10}}’}, {{V_{11}}’}, {{V_{12}}’}, {{V_{13}}’}, {{V_{14}}’}$\end{minipage}};
    
    \draw[thick] (8) -- (10) node[midway, color=blue, fill=white, swap] 
        {\begin{minipage}{3cm}\centering \small $V_1, V_3, V_4, V_6, V_7,$\\$ V_8, {{V_{10}}’}, {{V_{11}}’}, {{V_{12}}'}, {{V_{13}}’}, {{V_{14}}’}$\end{minipage}};
\draw[thick, bend left=45] (8) to node[midway, color=blue, fill=white] {\begin{minipage}{3cm}\centering \small $V_3, V_6, V_7,$\\${{V_{10}}’}, {{V_{11}}’}, {{V_{12}}’},$\\${{V_{13}}’}$\end{minipage}} (11);
                \draw[thick] (11) -- (12) node[midway, color=blue,fill=white, swap] {\small$V_{7}, V'_{10}, V'_{11}, V'_{12}$};
                \draw[thick] (12) -- (13) node[midway, color=blue,fill=white, swap] {\small$V'_{11}, V'_{12}$};

             %   \draw[thick] (1) edge  (2);
             %  \draw[thick] (1) edge  (3);
             %   \draw[thick] (1) edge  (4);
              % \draw[thick] (1) edge[bend right=45]  (8);

             %\draw[thick] (2) edge  (5);
              % \draw[thick] (2) edge  (6);
               %     \draw[thick] (2) edge  (7);
               %     \draw[thick] (2) edge[bend right=25]  (8);    

               %   \draw[thick] (8) edge  (9);
               %     \draw[thick] (8) edge  (10);
               %     \draw[thick] (8) edge[bend left=35]  (11);                           

        \end{tikzpicture}}  	\caption{Cone-cloud's estimand's join-graph. 
    %\jin{?. This is not a dual graph. $F_{11}$ should not contain $V'_{14}$. }\rina{agree. Annie, please remove $V'_{14}$  from $F_{11}$}
    }
	\label{fig:cone_dual}
\end{figure*}}

\shrink{
\begin{figure}[t]
        \captionsetup{justification=centering}
\label{fig:coneHT}
  \begin{minipage}[b]{.85\columnwidth}
     \caption*{\scriptsize Key for Eqn.\ref{eq:cone_flattened}}
     \vspace{-2mm}
    \scriptsize    
      \resizebox{\linewidth}{!}{%
    \begin{tabular}{  c c  }%
    \toprule
     Name &  Fcn Scope\\
     \midrule
 $F_1$ & $V_2, V_4, V_5, V_7, V_8, V_9, V_{11}, V_{12}, V_{13}, V_{14}$ \\
$F_2$ & $V_9, V_{13}, V_{14}$ \\
$F_3$ & $V_8, V_{12}, V_{13}$ \\
$F_4$ & $V_1, V_3, V_4, V_6, V_7, V_8, V_{10}, V_{11}, V_{12}, V_{13}$ \\
$F_5$ & $V_7, V_{11}, V_{12}$ \\
$F_6$ & $V_6, V_{10}, V_{11}$ \\
$F_7$ & $V_{11}, V_{12}, V_{13}$ \\
$F_8$ & $V_0, V_1, V_2, V_3, V_4, V_5, V_6, V_7, V_8, V_9, V_{10}', V'_{11}, V'_{12}, V'_{13}, V_{14}'$ \\
$F_9$ &  $V_3, V_6, V_7, V_{10}', V'_{11}, V'_{12}, V'_{13}$ \\
$F_{10}$ & $V'_{11}, V'_{12}$ \\
$F_{11}$ & $ V_1, V_3, V_4, V_5, V_6, V_7, V_8, V_9, V_{10}', V'_{11}, V'_{12}, V'_{13}, V_{14}'$  \\
$F_{12}$ & $ V_1, V_3, V_4, V_6, V_7, V_8, V_{10}', V'_{11}, V'_{12}, V'_{13}, V_{14}'$ \\
$F_{13}$ & $V'_{10}, V_7, V'_{11}, V'_{12}$ \\  
    \bottomrule
    \end{tabular}}
    \label{tab:key_chain}
  \end{minipage}\\
\begin{subfigure}[t]{\columnwidth}
    \resizebox{.75\linewidth}{!}{\begin{tikzpicture}[-,>=stealth',shorten >=1pt,auto,node distance=6cm,
  main node/.style={ellipse,draw,font=\sffamily\small},
    every label/.style={font=\sffamily\small}]
        
            \node[main node] (1) at(0,0) { $F_1, F_4$};
            %children of node 1
            \node[main node] (2)at (-3,0) { $F_7$};
            \node[main node] (3) at (-2,-1) { $F_6$};
            \node[main node] (4)at (-.5, -2) {$F_8$};
            
            \node[main node] (5)at (1,-2) {$F_5$};
            \node[main node] (6) at (2.5,-1.25) {$F_3$};
            \node[main node] (7) at ( 3,0) {$F_2$};

            %children of node 8 $
            \node[main node] (8) at ( -3,-2) {$F_{11}$};
            \node[main node] (9) at ( -2,-3) {$F_{12}$};
            \node[main node] (10) at ( .75,-3) {$F_{9}$};
            \node[main node] (11) at ( 2.25,-3) {$F_{12}$};
            \node[main node] (12) at ( 4,-3) {$F_{10}$};

                \draw[thick] (1) -- (2) ;
                \draw[thick] (1) -- (3) ;
                \draw[thick] (1) -- (4) ;   
                \draw[thick] (1) -- (5) ;     
                \draw[thick] (1) -- (6);     
                \draw[thick] (1) -- (7) ;

                \draw[thick] (4) -- (8) ;
            
            \draw[thick] (4) -- (9) ;
       
                \draw[thick] (4) -- (10) ;  
                \draw[thick] (10) -- (11) ;  
                \draw[thick] (11) -- (12) ;  
        \end{tikzpicture}
      %  $ V_3,V_4,V_6, V_7, V_8, V'_{10}, V'_{11}, V'_{12}, V'_{13}$}
}  
\end{subfigure}
    \caption{Hypergraph for Eqn
\ref{eq:cone_flattened} with hyperwidth 2 and largest cluster containing 14 variables from $F_1$ and $F_4$}
    \label{fig:coneHT}%
\end{figure}

}

%\begin{figure}[htbp]
%	\centering
%	\includegraphics[scale=0.5]{figures/cone4-ex.pdf}
%	\caption{Cone estimand's dual graph.}
%	\label{fig:cone_dual}
%\end{figure}

\shrink{
We can also consider only the 3 maximal clusters (all others can be absorbed in them in linear time, yielding 3 clusters or hyper-edges:\\
$C_1 = \{ 2,4,5,7,8,9,11,12,13,14\}$ \\
$C_2 = \{ 1,3,4,6,7,10,11,12,13  \}$ \\
$C_1 = \{ 0,1,2,3,4,5,6,7,8,9,10',11',12',13',14' \}$\\
}
\end{example}
\begin{figure}[t]
  \begin{subfigure}{.33\columnwidth}
         \resizebox{\linewidth}{!}{\begin{tikzpicture}[->,>=stealth',shorten >=1pt,auto,node distance=1.5cm,
  main node/.style={circle,draw,font=\sffamily\small}]

  \node[main node] (1) {W};
  \node[main node] (2) [below left of=1] {R};
  \node[main node, color=blue] (3) [below left of =2] {X};
  \node[main node, color=red] (4) [right of=3] {Y};
  
   \draw[thick] (1) edge  (2);
  \draw[thick, dashed] (1) edge[bend right=25] (3);
    \draw[thick, dashed] (3) edge[bend left=25] (1);

    \draw[thick] (2) edge  (3);
    \draw[thick] (3) edge  (4);

  \draw[thick, dashed] (1) edge[bend left=25] (4);
    \draw[thick, dashed] (4) edge[bend right=25] (1);

\end{tikzpicture}} 
        \caption{Causal Graph}
         \label{fig:napkin}%
      \end{subfigure} 
  \hfill
        \begin{subfigure}{.6\columnwidth}
         \resizebox{\linewidth}{!}{\begin{tikzpicture}[-,>=stealth',shorten >=1pt,auto,node distance=2cm,
  main node/.style={ellipse,draw,minimum size=.25cm,font=\sffamily\small}]
        
    \node[main node, fill=blue!20] (1) at(1.25,0) {$X,Y,R,W$};
            \node[above=0.1cm of 1,xshift=0.5cm, blue] {\scriptsize$F( X,Y,R,W)$}; 
            
    \node[main node, fill=blue!20] (2) at (-1, 0) {$W$};
            \node[above=0.1cm of 2,xshift=0cm, blue] {\scriptsize$F( W)$}; 
    \node[main node, fill=blue!20] (3) at (4,0) {$X,R$};
            \node[above=0.1cm of 3,xshift=0cm, blue] {\scriptsize$g(X,R)$}; 
            
    \node[main node, fill=red!30] (4) at (-1,-1.5) {$W'$};
            \node[below=0.1cm of 4,xshift=0cm, red] {\scriptsize$F(W')$}; 
    \node[main node, fill=red!30] (5)at (1.25,-1.5){$X,R,W'$};  
            \node[below=0.1cm of 5,xshift=0cm, red] {\scriptsize$F(X,R,W')$}; 
    \node[main node, fill=red!30] (7)at (4,-1.5){$X,R$};  
            \node[below=0.1cm of 7,xshift=0cm, red] {\scriptsize$O_1(X,R)$}; 
           %\node (6) at(4,-2) {};
    \node[main node, fill = blue!20](8) at(0,1.2) {$X,Y$};

            \node[above=0.1cm of 8,xshift=0cm, blue] {\scriptsize$O_2(X,Y)$};

                \draw[thick] (1) -- (2) ;
                \draw[thick] (1) -- (3) ;
                \draw[thick] (4) -- (5) ;
                \draw[thick] (5) -- (7) ;
                
                %\path (4) -- (5) node[midway] (6) {};
                \draw[thick] (1) -- (8) ;

                \draw[thick, purple] (7) -> (3) ;

               \draw[thick, purple, ->] (7) to (3);

        \end{tikzpicture}} 
        \caption{Heirarchy Hypergraph}
         \label{fig:napkin_dual}%
      \end{subfigure} 
      \caption{Napkin Model}
      \vspace{-5mm}
\end{figure}
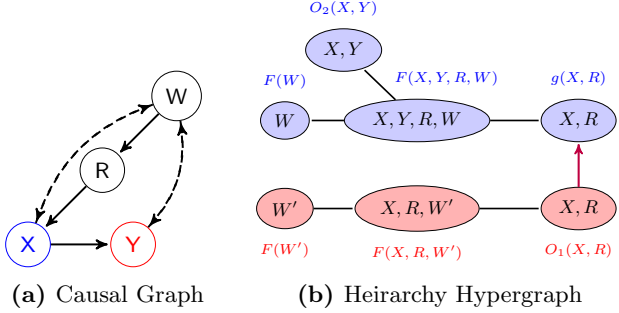

\paragraph{Ratio estimands.} It is sometime the case that estimands include ratios of expressions. How should we treat those? One option is to treat a denominator expression as a distinct sum-product expression which connects to its numerator by its ``output function''. This suggests a {\it sub-expressions hierarchy} which  dictates a sum-product processing order. The components of the hierarchy are flattened sum-product sub-expressions as seen in the previous examples, where each corresponds to a sum-product expression over a probabilistic graphical model. Each layer in the hierarchy is a single expression and its sum-product output function is processed by $CTE$ over its own hypertree decomposition. These output functions will connect the layers in a chain heirarchy, where the output function from each layer is included in the sub-expression of its parent layer.
We show the details in the next example followed by the general evaluation algorithm.

% We will illustrate with the following example.

%\vspace{-5mm}
\begin{example}\label{ratio-ex}
Consider the ratio estimand expression 
associated with the Napkin model of Figure \ref{fig:napkin}.
\[
P(Y|do(X))=\frac{\sum_W P(X,Y |R,W)P(W)}{\sum_{W'} P(X |R,W')P(W')}
\]
The denominator expression is composed of 2 functions whose scopes are the two sets $W'$ and $X, R, W'$. The corresponding dual graph is depicted underneath the red arrow in Figure \ref{fig:napkin_dual}. 
  %$W$ was renamed $W'$ for this sub expression. 
  The output function for this sum-product is a new function $O_1(X,R)$   whose inverse $g(X,R) = \frac{1}{O_1(X,R)}$ will be noted as a factor of its numerator. 
  %\annie{the following sentence is confusing, do you mean "The arrow emanating from node $O_1(X,R)$ to node $g(X,R)$ indicates that the function, $g(X,R)$, will be generated by the sum-prodcut in the previous level of the hierarchy" } 
  %\rina{yes.} 
  The red arrow emanating from node $O_1$ with arguments $\{X,R\}$ indicates that the function will be generated in that child layer and will be used in the parent layer.
  %child sum-product in the hierarchy. 
  The sum-product at the top layer of the hierarchy has 4 nodes corresponding to 4 functions; 2 in the numerator, one is the function $g(X,R)$ generated by the child expression, and $O_2$ is the output function produced as a result of the summation over $W$. Since, the denominator expression has $hw=1$, the tightness of the output function $g(X,R)$ remains bounded by $t^{1}=t$ (the data size).
 %\annie{why is t=D? we saw multiple examples where it can be far lower, $t \in O(D)$} \rina{it is just a bound}. 
 %\annie{I know but this is confusing, a constant interchange between t and d as if they are equal, I would wonder then why the bound is written in terms of $t$ and not $d$ if im reading this and not familiar }\rina{what is d? in any case the theory is on Bayesian network where the tightness is the number of tuples in the functions. There is no data there.} \annie{i know thats my point...}
 Since the numerator's expression including the output function $g(X,R)$ also has $hw=1$, the overall hyperwidth of the hierarchy of this estimand is $hw=1$ and since the tightness does not change, the computation would be linear in the tightness which is bounded by the data size.
\end{example}

%The flattened sum-products hierarchy is a directed path. 
Due to the flattening process each sum-product expression will have a single sum-product child from a denominator (see the hierarchy graph in Figure \ref{fig:napkin_dual}). We next provide more details.

%\begin{definition}[output function]
%The function expressing the answer for a flat sum-product expression is called it {\it output function}
%\end{definition}

%\shrink{
\begin{definition}[sum-product chain hierarchy]
Given an estimand whose  expressions are all flattened. The nodes of the sum-product hierarchy are flat sum-product expressions.
%\annie{root node sounds weird in this context, makes it sounds like one node, i suggest saying top layer}
The top layer is the numerator sum-product expression. Each denominator sum-product expression is a child of its direct numerator sum-product expression. 
The function generated by a sum-product computation is called its {\it output function}.
%\rina{should this coming paragraph be removed?}\annie{yes}
\shrink{
Visually We can associate each sum-product expression in the hierarchy with its dual graph or hypergraph representation treating the expression as a Bayesian network (see definition \ref{def3}) which creates a graphical hierarchy that corresponds to the sub-expression hierarchy. Each level includes an additional node for the output function of the sum-product computed at this level. 
There is a  directed edges leading from the output function node of the child's dual graph to an input node in the parent dual graph having the same set of variables but the function is the inverse input node. See figure \ref{fig:napkin_dual} for illustration. We will show during the estimand's evaluation each dual graph is embedded in a hypertree creating a chain of hypertrees that guide the computation. 
}
\shrink{
Each sum-product hierarchy (a chain) can be associated with a {\bf dual graphs hierarchy}, 
where at each level we have the dual graph of the corresponding sum-product expression in the hierarchy. Each level includes an additional node for the output function it generates and for the output function its child expression generates.
there is a  directed edges leading from the output function node of the child's  dual graph to an input node in the parent dual graph having the same set of variables but the function is the inverse input node. See figure 5 for illustration.
}
\end{definition}
%}

%\begin{definition}[The heirarchy dual graph] 
%\annie{are we calling it a hypergraph or a dual graph or a hyper treee}
%The dual graph of the heirarchy is a directed path of dual graphs, each associated with a level in the hierarchy. Each sub-expression has a node for its output function. The output function node is duplicated and included also in the parent expression.
%\end{definition}

\paragraph{Algorithm \ref{alg:pluginhypertree}} describes our plug-in hypertree evaluation (PI-HTE) algorithm. Its input is an estimand and a dataset. The first step is flattening, where for each sum-product expression the summation variables are renamed in all the functions within the summation's scope. 
Step 2 creates the sum-product chain hieararchy as defined above. 
%It can be defined recursively, starting with the flat numerator expression as the root. Each node is a flat sum-product expression.
%This yield a directed path whose root is the primary numerator expression.
In steps 3-10 the algorithm processes the sum-product hierarchy in order from leaf to root. 
%Each sum-product level defines its {\it output function}  $O_i$ for level $i$ in the hierarchy. 
%Each sub-expression corresponds to an empirical Bayesian network for which a sum-product query needs to be evaluated  to create the output function of that layer (step 5).
When layer $i$ is processed, its empirical BN is well defined. Then a hypertree decomposition is created by a hypertree decomposition scheme, taking into account the output function obtained from its child level, to which $CTE$ can be applied producing its own output function $O_i$.
 The output function is then inverted (step 9) and incorporated into the dual graph of the parent, and computation continues with the parent layer. The answer is obtained in the output function of the root layer.

\begin{theorem}[soundness]
Algorithm {PI-HTE} is sound for estimand evaluation given a dataset.
\end{theorem}

\vspace{-3mm}
\paragraph{Complexity.}
We associate the dual graph of level $i$ with a hypertree decomposition having hyperwidth $hw_i$ and treewidth $w_i$. 
%This yield a chain of hypertrees that are connected via the output function nodes at each level. 
The complexity of computing the output function in each level is dominated by $t^{hw_i}$ and $k^{w_i}$.
Thus, the overall complexity is controlled by
the treewidths and hypertree widths over all the levels. This seems to imply that complexity is exponential in the largest $hw$ and the largest $w$; however, for hypertree width the complexity can be more involved.
%his, indeed will be the case if the output functions generated in the process will have the same tightness as our input functions in the expression.
The tightness of the output functions may no longer be bounded by the data size $t$.
Rather, the worst-case tightness of $O_i$ is $t^{hw_i}$.
%A more refined analysis can be done however. Let the cluster in which out(i) resides in the HTD of layer i be $C_i$. and let $l_i$ be its number of functions. It is easy to see that the tightness of $out(i)$ will be $t^{l_i}$ where $l_i$ can be far smaller than $hw_i$.  Moving to the parent layer, the output cluster will be associated with $t^{l_i}$ tightness. In the parent layer $i+1$ we can take into account the cluster that includes the $out(i)$ to derive a lower bound. In the following we state the complexity of the algorithm as a function of $hw$ of each layer without taking the refined description into account.
The following theorem summarizes.
Since the dominant factors in the complexity are $t^{hw}$, and $k^w$), for simplicity we will drop the $log t$ factor appearing in the complexity bound.

{\noindent
\begin{algorithm}[t!]
    %\caption{EM4CI($k_{hyp}$)}
    \caption{Plug-In Hypertree Evaluation (PI-HTE)}
    \label{alg:pluginhypertree}
    {\bf Input:} data $\D$, and estimand expression for $P({\bf Y} \mid do({\bf X}))$ over a given DAG $G$.\\
    {\bf Output:} an estimate of $P({\bf Y} \mid do({\bf X}))$
    \hrule
    \vspace{6pt}

%   \SetAlgoHangIndent{\widthof{Step 1:~~~~}}

    1.\hspace{0.1cm}
{\bf Flatten sum-products} by renaming and moving\\ \makebox[.5cm]{} all summations to the front. \\
    ~~2.\hspace{0.1cm}
        Generate a hierarchy of sum product expressions.\;\\
    %\vspace{2pt}\\
    ~~3.\hspace{0.1cm}  Let $P$ be the lowest sum-product expression.\\
   ~~4.\hspace{0.1cm} \textbf{While} $P$ has a parent expression \textbf{do} \\
    ~~5.~~~~ {Let $\G$ be the DAG defined by functions in P} \\
   ~~6.~~~~ {Generate  $empBN(\G,D)$} \\
 ~~7.~~~~ {Generate a hypertree $H_P$ over $empBN(P,\D)$}\\
 \makebox[.8cm]{} aiming for $\min(hw_P)$.\\
 ~~8.~~~~  $O_P\leftarrow CTE(H_P)$\\
 ~~9.~~~~ $g_P= \frac{1}{O_P}$; $parent(P)= parent(P)\cup \{g_P\}$\\
 ~~10.~~~~$P \leftarrow  parent(P)$ \\
    ~~11.~~{\bf EndWhile}\\
    ~~12:\hspace{0.1cm}
    Return ~~  $P({\bf Y} \mid do({\bf X}))$\; %(e.g., by BE) \;
    %\rina{return?}
\end{algorithm}
\vspace{-5mm}
}
\begin{theorem}[complexity] Given an estimand expression having sum-product  hierarchy of depth $l$, algorithm PI-HTE has time and space complexity
{$O(n \cdot t^{\sum_{i=1}^lhw_i})$}, where $n$ is the number of variables, $t$ the tightness, and $hw_i$ is the hypertree width of the expression at level $i$. The algorithm's complexity is also
$O(n \cdot k^{max_{i=1}^l\{(w_i)\}})$ where $w_i$ is the treewidth of the expression at level $i$.  
\end{theorem}

\begin{proof}
(proof by induction)
Case 1: It is clear that if we have no denominators the estimand has only one level.  This defines a sum-product product expression. Since it can be viewed as an empirical Bayesian network having tightness $t$ it can be processed in $O(n \cdot t^{hw})$ (Theorem \ref{thm3a}), or by $O(n \cdot t^{w+1})$.
This proves the base case of l=1.
%Lets assume we have l=2.
When $l>1$  the inductive hypothesis for level $i-1$ assumes that the effective hyper-width  for that level is  $\sum_{j=1}^{i-1}hw_j$. 
This implies that  
its output function,  $O_{i-1}$, computed at level $i-1$ has tightness bounded also by $t^{\sum_{j=1}^{i-1}hw_j}$.
%$O((t_{i-1})^{hw_{i-1}})$. 
The output function $O_{i-1}$ is placed  at the parent level $i$, and grouped in one of the clusters with at most $hw_{i}$ functions having tightness $t$ in that level. The product of $hw_i$ functions of tightness $t$ with the single $O_{i-1}$ function having tightness $t^{\sum_{j=1}^{i-1}hw_j}$
%$t_{i-1}^{hw_{i-1}}$  
yields computation of at most $O(t^{hw_i} \cdot t^{\sum_{j=1}^{i-1}hw_j} )= O(t^{\sum_{j=1}^i hw_j})$. Thus the effective hyperwidth grows additively with each level both time and memory, yielding the claim.
The bound based on treewidth is not dependant on the tightness but only on the domain size and we therefore get that the relevant treewidth that dominates the computations is the maximum treewidth over the whole hierarchy.
Note that when $O_{i-1}$ resides in its own cluster in the tree decomposition of its parent level $i$, the effective hyperwidth of level $i$ is
just the max between the hyperwidth of the child expression and its parent expression. If this occurs at each level the effective hyperwidth of the whole computation is the maximum hyperwidth over all levels. In this case the bound is $O(t^{\max_{j=1}^i hw_j})$.

\end{proof}

%\vspace{-10mm}
\begin{table}[h!]
       \centering
        \captionsetup{justification=centering}
        \caption{
        Empirical plug-in results for different sample sizes. Here $t$ is the tightness of the data, and $k$ is the variables' domain sizes.}
    \begin{subtable}{\columnwidth}
    \caption{\scriptsize Chain: $P(V_{99} | do(V_{0}))$\\ $|\V| =99$; $ ~|\U|=49$ hw $=1$, tw=98\\ $P(V_{98} |V0, \ldots V_{97} )$- largest factor }
    \scriptsize    
      \resizebox{\columnwidth}{!}{%
    \begin{tabular}{  c c c c c }%
    \toprule
    \textbf{\#Samples} & \textbf{time} & \textbf{Max table size} & \textbf{t} & \textbf{density} \\
     \midrule
  100 & 11.2 & 100 & 100 & $9.9 \times 10^{-34}$ \\
200 & 19.7 & 200  & 200 & $2.0 \times 10^{-33}$ \\
400 & 31.4 & 400  & 400 & $3.9 \times 10^{-33}$ \\
800 & 53.7 & 800  & 800 & $7.9 \times 10^{-33}$ \\
1,000 & 69.5 & 1,000  & 1,000 & $1.0 \times 10^{-32}$ \\
1,600 & 97.5 & 1,600  & 1,600 & $1.6 \times 10^{-32}$ \\
3,200 & 192.4 & 3,200  & 3,200 & $3.2 \times 10^{-32}$ \\
6,400 & 369.2 & 6,400  & 6,400 & $6.4 \times 10^{-32}$ \\
10,000 & 608.8 & 10,000  & 10,000 & $9.9 \times 10^{-32}$ \\
    \bottomrule
    \end{tabular}}
    \label{tab:99chain_plug_in}
\end{subtable}\\
\label{tab:exp-results}
    \begin{subtable}{\columnwidth}
        \centering
        \caption{\scriptsize Cone Cloud $P(V_0 | do(V_{14}, V_{10}, V_{4}))$\\ $|\V| =15$; $ ~|\U|=8$; $ k=10$; hw $=2$; $tw=14$\\
        largest factor $P(V_0 | V_1 \ldots V_{14})$}
            \scriptsize    
        \resizebox{\columnwidth}{!}{\begin{tabular}{ccccc}
            \toprule
            \textbf{\#Samples} & \textbf{time (s)} & \textbf{Max table size}  & \textbf{t}& \textbf{density}\\
            \midrule
    100    & 2.2    & 9,900          & 100  & 1e-13 \\
    200    & 3.3    & 39,203         & 200  & 2e-13 \\
    400    & 7.3    & 152,877       & 400  & 4e-13 \\
    800    & 28.8   & 589,815       & 800  & 8e-13 \\
    1,000   & 39.1   & 882,604       & 1,000 & 1e-12 \\
    1,600   & 84.8   & 2,206,528     & 1,600 & 1.6e-12 \\
    3,200   & 280.3  & 7,553,700     & 3,200 & 3.2e-12 \\
    6,400   & 788.74 & 21,949,125   & 6,400 & 6.4e-12 \\
    10,000  & 1,594.2 & 40,404,630   & 10,000 & 1e-11 \\
            \bottomrule
        \end{tabular}}
    \end{subtable}\\
     \begin{subtable}{\columnwidth}
        \centering
        \caption{\scriptsize Diamond Model, $P(V_{16}| do(V_0, V_4))$\\ $|\V| =17$;$k=10$; hw $=2$, tw=16\\$P(V_{16} |V_0, \ldots V_{15})$- largest factor}
            \scriptsize    
       \resizebox{\columnwidth}{!}{ \begin{tabular}{cccccc}
            \toprule
            \textbf{\#Samples} & \textbf{time (s)} & \textbf{Max table size} &  \textbf{t} &\textbf{density}\\
            \midrule
    100    & 0.6     & 8,672             & 100     & $1 \times 10^{-15}$  \\
200    & 1.9     & 35,605           & 200     & $2 \times 10^{-15}$  \\
400    & 6.9     & 156,160         & 400     & $4 \times 10^{-15}$  \\
800    & 26.4    & 603,790        & 800     & $8 \times 10^{-15}$  \\
1,000   & 39      & 928,510      & 1,000    & $1 \times 10^{-14}$  \\
1,600   & 87.4    & 2,184,880     & 1,600    & $1.6 \times 10^{-14}$\\
3,200   & 299.4   & 7,600,350     & 3,200    & $3.2 \times 10^{-14}$\\
6,400   & 988.8   & 22,208,750    & 6,400    & $6.4 \times 10^{-14}$\\
10,000 & 2,037.5  & 39,943,750   & 10,000  & $1 \times 10^{-13}$  \\

        \bottomrule

\end{tabular}}
    \label{tab:3layer1_plug_in}
    \end{subtable}\\
    \begin{subtable}{\columnwidth}
        \centering
        \caption{\scriptsize 3 Layer Model, $P(\textbf{V}| do(Z,T))$\\ $|\V| =17$; $2 \leq k \leq 50$; hw $=4$, tw=15\\ $P(V_0 |V_1, V_2, V_3, X_0)$- largest factor }
            \scriptsize    
       \resizebox{\columnwidth}{!}{ \begin{tabular}{ccccc}
            \toprule
            \textbf{\#Samples} & \textbf{time (s)} & \textbf{Max table size}  & \textbf{t} & \textbf{density} \\
            \midrule
25   & 2.8     & 3,219          & 25  & 0.00125  \\
50   & 26.5    & 79,136         & 50  & 0.0025   \\
100  & 640.8   & 1,623,559    & 100 & 0.005    \\
200  & 30,511.5 & 34,018,439  & 200 & 0.01     \\
\end{tabular}}
    \label{tab:3layer2_plug_in}
    \end{subtable}
    \vspace{-5mm}
\end{table}

%}
\shrink{
\begin{figure}[ht]
    \centering
    \caption{Comparative Plots for the Chain Model}
        \centering
        \includegraphics[width=\columnwidth]{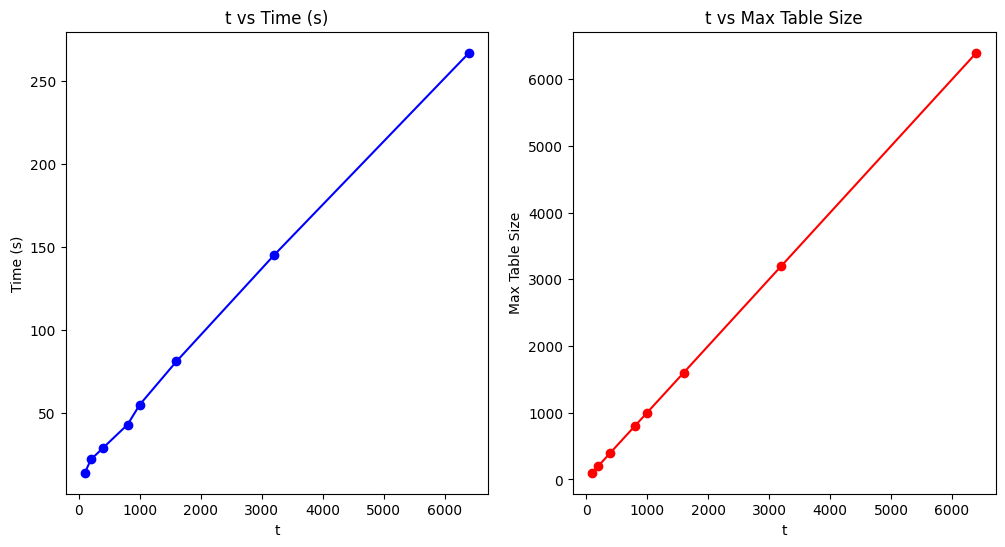}
    \label{fig:combined_plots}
\end{figure}
\shrink{
\begin{figure}[ht]
    \centering
    \begin{subfigure}[b]{0.45\columnwidth}
        \centering
        \includegraphics[width=\textwidth]{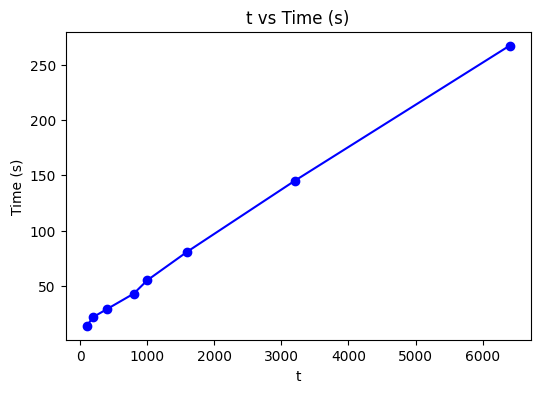}
        \caption{t vs Time (s)}
        \label{fig:t_vs_time}
    \end{subfigure}
    \hfill
    \begin{subfigure}[b]{0.45\columnwidth}
        \centering
        \includegraphics[width=\textwidth]{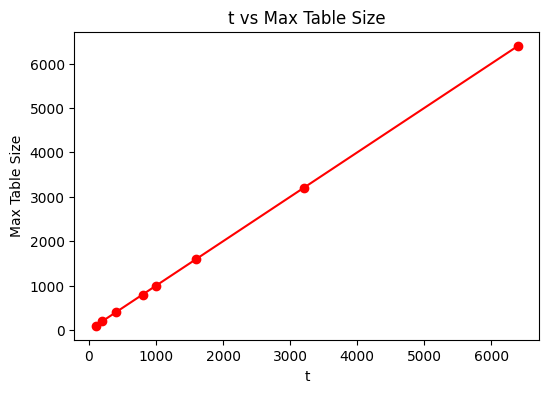}
        \caption{t vs Max Table Size}
        \label{fig:t_vs_max_table_size}
    \end{subfigure}
    \caption{Comparative Plots for the Chain Model}
    \label{fig:combined_plots}
\end{figure}
}}

%\shrink{
\section{Empirical validation}
\subsection{Benchmarks.}
For benchmarks we used three semi-markovian examples, and one markovian (without latent confounding). The semi-markovian examples were taken from \cite{DBLP:journals/corr/abs-2408-14101}, and the markovian example was constructed to have higher hyperwidth. Since, $O(t^{hw})$ is a worst case bound we ran our algorithm on various data sets, and present results that are close to worst case. We sampled data different distributions: uniform, deterministic, Dirichlet, and their mixes. Queries were chosen to correspond to complex estimands that could pose computational challenges.
%\annie{ my data sets were pretty hacky but I feel like this is an accurate portrayal of what I actually did.}
\subsection{Performance Measures}
We report the time and max table size during computation of the empirical plug-in on different sample sizes. The max table size reflects the space needed to evaluate the estimand. We also report the tightness $t$, and the $density$, which gives us a measure on the largest factor in the estimand, where ${density = \frac{\text{\# entries in table}}{\text{\# of configurations in the joint table }}}$.
\subsection{Results}

The empirical plug-in results with varying sample sizes are presented in Table \ref{tab:exp-results} on the Chain, Cone-cloud, Diamond, and
3-layer models (Figure \ref{fig:all_models}). We show that the complexity for estimand evaluation is reflected in the hyperwidth associated with the estimand, and it provides a much tighter bound than the treewidth.

The disparity between bounds is most evident in the Chain model where we have hyperwidth of one, thus computation is linear in the tightness; but the treewidth is 98, with max domain size $k=4$. So bounding the space and time complexity in the treewidth we get $O(k^{99}) \!\approx\! 4\!\times\!10^{59}$ for time and $O(k^{98}) \approx 1\!\times\! 10^{59}$ for space, giving us a loose and irrelevant bound. Even with 10,000 samples and $t=10,000$ the bound from the hyperwidth is $O(t^{1}) = O(10,000^{1})$, showing it's a more informative bound. 

We also show that the performance can be non-linear.  For example, in the case of the Cone-cloud and Diamond models, both with $hw=2$. %\rina{should we say how we know that this is the hw?. We show it for cones but not for diamonds or 3-layer. }\annie{i did the hypertree decomposition its just too big to include? I could say see supplmental and include it there, i did it piecewise as a hierarchy tho}
Here $t$ vs time and $t$ vs max table size are both $O(t^2)$. For example when $t=1,000$ the max table size for the cone-cloud and diamond, respectively, is 882,604 and 928,510, just under $1,000^2$. We also see that $t$ vs time grows faster than linearly, and is $O(t^2)$. Again, the treewidth is much higher than the hyperwidth, and the $max(k)$ is 10 and 50, so the treewidth bound is large and uninformative.
%3. In the 3 layers networks we  have hw=4 and tw =. This table illustrate how complexity scale up as a function of hw....???

In the 3-layer model we have hyperwidth 4 and treewidth 15, with a max domain size of 50. The relationship of $t$ vs time and $t$ vs max table size appears roughly cubic. So, while the worst-case is not realized we see significant impact of the hyperwidth.
%, and thus $ O(t^4)$ \rina{this is weird, even if not incorrect}\annie{so here it seems roughly cubic not as high as $t^4$ though, should I just state $O(t^4)$}. 
On the other hand, we do not see any values getting as high as the treewidth suggests ($O(50^{15})$). 
%We can see from that both time and space are far under this bound, and grow consistently in $O(t^4)$. 

%Thus, the hyperwidth, provides a tighter bound than is suggested from the tree width.
%This trend continues, for the other models and estimands, showing that the hyperwidth provides a tighter bound that is suggested from the treewidth for these estimands when we have sparse tables.

Since the tightness can increase with number of samples, the bound dictated by the hyperwidth can increase with number of samples. This is something to consider if you have a large number of samples with dense tables. 

Hypertree decompositions and estimands for the 3-layer and Diamond models are provided in the Supplemental.

\section{Conclusion}
The paper provides a new algorithm and analysis for evaluating plug-in estimands for causal effect queries.
The approach uses structural parameters, the treewidth and hypertree width, in bounding the complexity of plug-in estimand evaluation, similar to their role in graphical models schemes.
We show that the hypertree width is more informative for this task because of its sensitivity to function sparseness, and evaluating plug-in estimands from data yields sparse functions. 

We introduce a new algorithm Plug-in hypertree evaluation (PI-HTE) which harnesses hypertree decomposition algorithms 
%which exploit well known schems uses a message-passing scheme over hypertrees, 
to efficiently estimate the estimand, and we evaluate its performance over several classes of benchmarks and causal effect queries.
Our results confirm the significance of the hypertree parameter in capturing the algorithm's performance and its superiority over the treewidth parameter in the context of causal effect evaluation. In particular, it enables the evaluation of estimands previously thought to be computationally infeasible.

%In particular, the algorithm is time and space complexity is bounded by the hyperwidth, 
%and our experiments show the tightness of the hypertree width bound in practice using the empirical plug-in method. This enables the evaluation of estimands previously thought to be computationally difficult.

Our bounds help  characterize the computational feasibility of the empirical plug-in scheme, 
 establishing it as a simple and practical baseline for causal effect estimation.
 In particular, since  a causal query can have many candidate estimands, tree-width and hyperwidth-based bounds can be used as one metric to selecting among different estimands.

\clearpage

\bibliographystyle{natbib}
\bibliography{ref}
\clearpage
\section*{Checklist}

% %%% BEGIN INSTRUCTIONS %%%
%The checklist follows the references. For each question, choose your answer from the three possible options: Yes, No, Not Applicable.  You are encouraged to include a justification to your answer, either by referencing the appropriate section of your paper or providing a brief inline description (1-2 sentences). 
%Please do not modify the questions.  Note that the Checklist section does not count towards the page limit. Not including the checklist in the first submission won't result in desk rejection, although in such case we will ask you to upload it during the author response period and include it in camera ready (if accepted).

%\textbf{In your paper, please delete this instructions block and only keep the Checklist section heading above along with the questions/answers below.}
% %%% END INSTRUCTIONS %%%

 \begin{enumerate}

 \item For all models and algorithms presented, check if you include:
 \begin{enumerate}
   \item A clear description of the mathematical setting, assumptions, algorithm, and/or model. \\yes
   \item An analysis of the properties and complexity (time, space, sample size) of any algorithm. \\
   Yes in section 2 and section 3
   \item (Optional) Anonymized source code, with specification of all dependencies, including external libraries. [Yes/No/Not Applicable]
 \end{enumerate}

 \item For any theoretical claim, check if you include:
 \begin{enumerate}
   \item Statements of the full set of assumptions of all theoretical results.\\
   yes, we explain the setting for causal effect estimation in section 2
   \item Complete proofs of all theoretical results.\\
   yes
   \item Clear explanations of any assumptions. \\
    yes
\end{enumerate}

 \item For all figures and tables that present empirical results, check if you include:
 \begin{enumerate}
   \item The code, data, and instructions needed to reproduce the main experimental results (either in the supplemental material or as a URL). \\
    yes, can include in supplemental
   \item All the training details (e.g., data splits, hyperparameters, how they were chosen). \\not applicable
    \item A clear definition of the specific measure or statistics and error bars (e.g., with respect to the random seed after running experiments multiple times). \\
    not applicable
    \item A description of the computing infrastructure used. (e.g., type of GPUs, internal cluster, or cloud provider).\\
    no, we are showing worst case results, and interested in only how the results for in $O(t^{hw})$
 \end{enumerate}

 \item If you are using existing assets (e.g., code, data, models) or curating/releasing new assets, check if you include:
 \begin{enumerate}
   \item Citations of the creator If your work uses existing assets. \\
   not applicable
   \item The license information of the assets, if applicable. \\
      not applicable
   \item New assets either in the supplemental material or as a URL, if applicable.\\
     not applicable
   \item Information about consent from data providers/curators.\\
   not applicable
   \item Discussion of sensible content if applicable, e.g., personally identifiable information or offensive content. \\
   not applicable
 \end{enumerate}

 \item If you used crowdsourcing or conducted research with human subjects, check if you include:
 \begin{enumerate}
   \item The full text of instructions given to participants and screenshots. \\
   not applicable
   \item Descriptions of potential participant risks, with links to Institutional Review Board (IRB) approvals if applicable. \\
   not applicable
   \item The estimated hourly wage paid to participants and the total amount spent on participant compensation. \\
   not applicable
 \end{enumerate}

 \end{enumerate}

\clearpage

\shrink{

\appendix

\begin{figure}[t]
\caption{Hierarchy hypergraph for Eqn
\ref{eq:3layer_1}}
  \begin{minipage}[b]{.39\columnwidth}
     \caption*{\scriptsize Key for Eqn.\ref{eq:3layer_1}}
    \scriptsize    
      \resizebox{\linewidth}{!}{%
    \begin{tabular}{  c c  }%
    \toprule
     Name &  Fcn Scope\\
     \midrule
    $F_1$  & $X_0, T$   \\
    $F_2$  & $X_1, T$ \\
    $F_3$  & $X_2, T$     \\
    $F_4$  & $X_0, V_0, V_1, V_2, V_3$    \\
    $F_5$ & $X_0, V_3, V_4, V_5, V_6$   \\
    $F_6$ & $X_0, V_6, V_7, V_8, V_9$    \\
    $F_7$ & $X_0, V_9, V_{10}, V_{11}$  \\
    $F_8$ & $X_1, V_1$ \\ 
    $F_9$ &  $X_1, V_4$  \\
    $F_{10}$ & $X_1, V_7$ \\
    $F_{11}$ & $X_1, V_{10}$ \\
    $F_{12}$ & $X_2, V_2$ \\
    $F_{13}$ & $X_2, V_5$ \\
    $F_{14}$ & $X_2, V_8$ \\
    $F_{15}$ & $X_2, V_{11}$ \\
    $F_{16}$ & $Z, V_{0}, \ldots, V_{11} $ \\
    $F_{17}$ & $T$ \\
    $O_1$ &    $T,V_0, V_1, \ldots, V_{11}$    \\ 
    $O_2$ &       $T,Z$  \\
    \bottomrule
    \end{tabular}}
    \label{tab:key}
  \end{minipage}
\begin{subfigure}[t]{.59\columnwidth}
    \resizebox{\linewidth}{!}{\begin{tikzpicture}[-,>=stealth',shorten >=1pt,auto,node distance=3cm,
  main node/.style={ellipse,draw,minimum size=.25cm,font=\sffamily\small}]
        
            \node[main node, fill=blue!20] (1) at(0,.25) {$F_{16}, F_{17}$};
            \node[main node, fill=blue!20] (2) at (1, -1.5) {$O_2$};
            \node[main node, fill=blue!20] (3) at (-1,-1.5) {$O_1$};
            \node[main node, fill=red!30] (4) at (0,-4) {$F_{8},F_{9}, F_{10}, F_{11}$};
            \node[main node, fill=red!30] (5) at (2,-3) {$F_{12},F_{13}, F_{14}, F_{15}$};
            \node[main node, fill=red!30] (6)at (0,-6){$F_{1},F_{2},F_{3}, O_1$};  
            \node[main node, fill=red!30] (7)at (0,-8){$F_{4},F_{5},F_{6},F_{7}$};

             \draw[thick] (7) -- (6) node[midway, color=blue,fill=white] {\small$V_0, \ldots, V_{11}, X_0$};
               
             \draw[thick] (6) -- (4) node[midway, color=blue,fill=white] {\small$X_1,V_1, V_4,V_7, V_{10}$};

 \draw[thick, bend right=35] (6) to node[midway, color=blue, fill=white, swap] {\small$X_2, V_2, V_5, V_8, V_{11}$} (5);

 \draw[thick, purple, bend left=80] (6) to node[pos=.7, color=blue, fill=white, swap] {\small$V_0, \ldots, V_{11}, T$} (3);
 %              \draw[thick, purple] (6) -> (3) node[midway, color=blue,fill=white, swap] {\small$V_0, \ldots, V_{11}, T$};

              \draw[thick, purple,bend left=80, ->] (6) to (3);

        \draw[thick] (1) -- (2) node[midway, color=blue,fill=white] {\small$Z,T$};

    \draw[thick] (1) -- (3) node[midway, color=blue,fill=white,swap] {\small$V_0, \ldots, V_{11}, T$};

        \end{tikzpicture}} 
\end{subfigure}
    \label{fig:hh_3layer}%
\end{figure}

\begin{figure*}[!b]
	\centering
    \resizebox{.7\linewidth}{!}{\begin{tikzpicture}[-,>=stealth',shorten >=1pt,auto,node distance=6cm,
  main node/.style={ellipse,draw,minimum size=1.5cm,font=\sffamily\Large},
    every label/.style={font=\sffamily\Large}]
        
            \node[main node,text width=4cm, label=above:{$ F_1$}] (1) { $V_2, V_4, V_5,V_7,$$V_8,$\\$ V_9, V_{11},$$V_{12}, V_{13}, V_{14}$};  
            \node[main node,text width=3.5cm, label=above left:{$\large F_2$}] (2)[below left of =1] { $V_1, V_3, V_4, V_6, V_7,$\\$ V_8, V_{10}, V_{11}, V_{12}, V_{13}$};
            \node[main node, label=above right:{$F_3$}] (3)[right of =2] { $V_8, V_{12}, V_{13}$};
            \node[main node, label=above right:{$F_4$}] (4)[right of=3]{$V_9, V_{13}, V_{14}$};
            
            \node[main node,label=below right:{$F_5$}] (5)[below left of=2]{$V_7, V_{11}, V_{12}$};
            \node[main node, label=below right:{$F_6$}] (6)[right of=5]{$V_6, V_{10}, V_{11}$};
            \node[main node, label=below right:{$F_7$}] (7)[right of=6]{$V_{11}, V_{12}, V_{13}$};

            \node[main node, label=left:{$F_8$}] (8)[below left of =5,text width=3.5cm] { $V_0,V_1, V_2, V_3, V_4,$\\$ V_5, V_6, V_7, V_8, V_9, {{V_{10}}’}$\\$, {{V_{11}}’},{{V_{12}}’}, {{V_{13}}’}, {{V_{14}}’}$};
            \node[main node, label=below left:{$F_9$}] (9)[below left of =8, text width=3.5cm] { $V_1, V_3, V_4, V_5,$\\$ V_6, V_7, V_8, V_9,{{V_{10}}’},$\\$ {{V_{11}}’}, {{V_{12}}’}, {{V_{13}}’}, {{V_{14}}’}$};
            \node[main node, label=below right:{$F_{10}$}] (10)[right of =9, text width=4cm] { $V_1,  V_3, V_4, V_6, V_7, V_8,$\\$ {{V_{10}}’}, {{V_{11}}’}, {{V_{12}}’},{{V_{13}}’}, {{V_{14}}’}$};          
  \node[main node, label=below right:{$F_{11}$}] (11)[ right of=10, text width=3cm]{$V_3, V_6, V_7, {{V_{10}}’}, $\\${{V_{11}}’}, {{V_{12}}’}, {{V_{13}}’}$};
  \node[main node, label=below right:{$F_{12}$}] (12)[above right of=11]{$V_{7}, V'_{10}, V'_{11}, V'_{12}$};
    \node[main node, label=below right:{$F_{13}$}] (13)[ right of=12]{$V'_{11}, V'_{12},$};

                \draw[thick] (1) -- (2) node[midway, color=blue,fill=white, swap] {\small$V_4, V_7, V_8, V_{11}, V_{12}, V_{13}$};
                \draw[thick] (1) -- (3) node[midway, color=blue,fill=white, swap] {\small$V_8, V_{12}, V_{13}$};
                \draw[thick] (1) -- (4) node[midway, color=blue,fill=white, swap] {\small$V_9, V_{13}, V_{14}$};
\draw[thick, bend right=45] (1) to node[midway, color=blue, fill=white, swap] {\small$V_2, V_4, V_5, V_7, V_8$} (8);
                
                \draw[thick] (2) -- (5) node[midway, color=blue,fill=white, swap] {\small$V_7, V_{11}, V_{12}$};
                \draw[thick] (2) -- (6) node[midway, color=blue,fill=white, swap] {\small$V_6, V_{10}, V_{11}$};
                \draw[thick] (2) -- (7) node[midway, color=blue,fill=white, swap] {\small$V_{11}, V_{12}, V_{13}$};
    
\draw[thick, bend right=25] (2) to node [midway, color=blue, fill=white, swap] {\small$V_1, V_3, V_4, V_6, V_7, V_8$} (8);
                
 \draw[thick] (8) -- (9) node[midway, color=blue, fill=white, swap] 
        {\begin{minipage}{3cm}\centering \small $V_1, V_2, V_3, V_4, V_5,$\\$ V_6, V_7, V_8, V_9,$\\$ {{V_{10}}’}, {{V_{11}}’}, {{V_{12}}’}, {{V_{13}}’}, {{V_{14}}’}$\end{minipage}};
    
    \draw[thick] (8) -- (10) node[midway, color=blue, fill=white, swap] 
        {\begin{minipage}{3cm}\centering \small $V_1, V_3, V_4, V_6, V_7,$\\$ V_8, {{V_{10}}’}, {{V_{11}}’}, {{V_{12}}'}, {{V_{13}}’}, {{V_{14}}’}$\end{minipage}};
\draw[thick, bend left=45] (8) to node[midway, color=blue, fill=white] {\begin{minipage}{3cm}\centering \small $V_3, V_6, V_7,$\\${{V_{10}}’}, {{V_{11}}’}, {{V_{12}}’},$\\${{V_{13}}’}$\end{minipage}} (11);
                \draw[thick] (11) -- (12) node[midway, color=blue,fill=white, swap] {\small$V_{7}, V'_{10}, V'_{11}, V'_{12}$};
                \draw[thick] (12) -- (13) node[midway, color=blue,fill=white, swap] {\small$V'_{11}, V'_{12}$};

             %   \draw[thick] (1) edge  (2);
             %  \draw[thick] (1) edge  (3);
             %   \draw[thick] (1) edge  (4);
              % \draw[thick] (1) edge[bend right=45]  (8);

             %\draw[thick] (2) edge  (5);
              % \draw[thick] (2) edge  (6);
               %     \draw[thick] (2) edge  (7);
               %     \draw[thick] (2) edge[bend right=25]  (8);    

               %   \draw[thick] (8) edge  (9);
               %     \draw[thick] (8) edge  (10);
               %     \draw[thick] (8) edge[bend left=35]  (11);                           

        \end{tikzpicture}}  	\caption{Cone-cloud's estimand's join-graph. 
    %\jin{?. This is not a dual graph. $F_{11}$ should not contain $V'_{14}$. }\rina{agree. Annie, please remove $V'_{14}$  from $F_{11}$}
    }
	\label{fig:cone_dual}
\end{figure*}
\begin{figure}
    \centering
    \resizebox{\linewidth}{!}{\begin{tikzpicture}[-,>=stealth',shorten >=1pt,auto,node distance=3.5cm,
  main node/.style={ellipse,draw,minimum size=1cm,font=\sffamily\small}]
        
            \node[main node] (1) {$V_0, V_1, V_2, V_3, V_4, V_5$};
            \node[main node] (2)[below left of =1] {$V_0, V_1, V_2, V_3$};
            \node[main node] (3)[right of=2] {$V_0, V_1$};
            \node[main node] (4)[right of=3, text width=2cm]{${{V_0}'}, V_1, V_2, V_3,$\\$V_4, V_5, V_6$};
            \node[main node] (5)[below left of=4]{${{V_0}’}, V_1, V_2, V_3, V_4$};    
            \node[main node] (6)[below right of=4]{${{V_0}’}, V_1, V_2$};           
            \node[main node] (7)[right of=6]{${{V_0}’}$};

                \draw[thick] (1) -- (2) node[midway, color=blue,fill=white, swap] {\small$V_0, V_1, V_2, V_3$};
                \draw[thick] (1) -- (3) node[midway,color=blue, fill=white] {\small$V_0, V_1$};
                \draw[thick] (1) -- (4) node[midway, color=blue,fill=white] {\small$V_1, V_2, V_3, V_4, V_5$};
                \draw[thick] (4) -- (5) node[midway, color=blue, fill=white] {\small${{V_0}'}, V_1, V_2, V_3, V_4$};
                \draw[thick] (4) -- (6) node[midway, color=blue, fill=white] {\small${{V_0}'}, V_1, V_2$};
                \draw[thick] (6) -- (7) node[midway,color=blue, fill=white] {\small${{V_0}'}$};

                %\draw[thick] (1) edge  (2);
               %\draw[thick] (1) edge  (3);
               % \draw[thick] (1) edge  (4);
               %\draw[thick] (4) edge  (5);
             %\draw[thick] (4) edge  (6);
              %  \draw[thick] (6) edge  (7);

        \end{tikzpicture}}  
    \caption{}
    \label{fig:enter-label}
\end{figure}

\begin{figure}
    \centering
    \resizebox{\linewidth}{!}{\begin{tikzpicture}[-,>=stealth',shorten >=1pt,auto,node distance=6cm,
  main node/.style={ellipse,draw,minimum size=1.5cm,font=\sffamily\Large},
    every label/.style={font=\sffamily\Large}]
        
            \node[main node,text width=4cm, label=above:{$ F_1$}] (1) { $V_2, V_4, V_5,V_7,$$V_8,$\\$ V_9, V_{11},$$V_{12}, V_{13}, V_{14}$};  
            \node[main node,text width=3.5cm, label=above left:{$\large F_2$}] (2)[below left of =1] { $V_1, V_3, V_4, V_6, V_7,$\\$ V_8, V_{10}, V_{11}, V_{12}, V_{13}$};
            \node[main node, label=above right:{$F_3$}] (3)[right of =2] { $V_8, V_{12}, V_{13}$};
            \node[main node, label=above right:{$F_4$}] (4)[right of=3]{$V_9, V_{13}, V_{14}$};
            
            \node[main node,label=below right:{$F_5$}] (5)[below left of=2]{$V_7, V_{11}, V_{12}$};
            \node[main node, label=below right:{$F_6$}] (6)[right of=5]{$V_6, V_{10}, V_{11}$};
            \node[main node, label=below right:{$F_7$}] (7)[right of=6]{$V_{11}, V_{12}, V_{13}$};

            \node[main node, label=left:{$F_8$}] (8)[below left of =5,text width=3.5cm] { $V_0,V_1, V_2, V_3, V_4,$\\$ V_5, V_6, V_7, V_8, V_9, {{V_{10}}’}$\\$, {{V_{11}}’},{{V_{12}}’}, {{V_{13}}’}, {{V_{14}}’}$};
            \node[main node, label=below left:{$F_9$}] (9)[below left of =8, text width=3.5cm] { $V_1, V_3, V_4, V_5,$\\$ V_6, V_7, V_8, V_9,{{V_{10}}’},$\\$ {{V_{11}}’}, {{V_{12}}’}, {{V_{13}}’}, {{V_{14}}’}$};
            \node[main node, label=below right:{$F_{10}$}] (10)[right of =9, text width=4cm] { $V_1,  V_3, V_4, V_6, V_7, V_8,$\\$ {{V_{10}}’}, {{V_{11}}’}, {{V_{12}}’},{{V_{13}}’}, {{V_{14}}’}$};          
  \node[main node, label=below right:{$F_{11}$}] (11)[ right of=10, text width=3cm]{$V_3, V_6, V_7, {{V_{10}}’}, $\\${{V_{11}}’}, {{V_{12}}’}, {{V_{13}}’}$};
  \node[main node, label=below right:{$F_{12}$}] (12)[above right of=11]{$V_{7}, V'_{10}, V'_{11}, V'_{12}$};
    \node[main node, label=below right:{$F_{13}$}] (13)[ right of=12]{$V'_{11}, V'_{12},$};

                \draw[thick] (1) -- (2) node[midway, color=blue,fill=white, swap] {\small$V_4, V_7, V_8, V_{11}, V_{12}, V_{13}$};
                \draw[thick] (1) -- (3) node[midway, color=blue,fill=white, swap] {\small$V_8, V_{12}, V_{13}$};
                \draw[thick] (1) -- (4) node[midway, color=blue,fill=white, swap] {\small$V_9, V_{13}, V_{14}$};
\draw[thick, bend right=45] (1) to node[midway, color=blue, fill=white, swap] {\small$V_2, V_4, V_5, V_7, V_8$} (8);
                
                \draw[thick] (2) -- (5) node[midway, color=blue,fill=white, swap] {\small$V_7, V_{11}, V_{12}$};
                \draw[thick] (2) -- (6) node[midway, color=blue,fill=white, swap] {\small$V_6, V_{10}, V_{11}$};
                \draw[thick] (2) -- (7) node[midway, color=blue,fill=white, swap] {\small$V_{11}, V_{12}, V_{13}$};
    
\draw[thick, bend right=25] (2) to node [midway, color=blue, fill=white, swap] {\small$V_1, V_3, V_4, V_6, V_7, V_8$} (8);
                
 \draw[thick] (8) -- (9) node[midway, color=blue, fill=white, swap] 
        {\begin{minipage}{3cm}\centering \small $V_1, V_2, V_3, V_4, V_5,$\\$ V_6, V_7, V_8, V_9,$\\$ {{V_{10}}’}, {{V_{11}}’}, {{V_{12}}’}, {{V_{13}}’}, {{V_{14}}’}$\end{minipage}};
    
    \draw[thick] (8) -- (10) node[midway, color=blue, fill=white, swap] 
        {\begin{minipage}{3cm}\centering \small $V_1, V_3, V_4, V_6, V_7,$\\$ V_8, {{V_{10}}’}, {{V_{11}}’}, {{V_{12}}'}, {{V_{13}}’}, {{V_{14}}’}$\end{minipage}};
\draw[thick, bend left=45] (8) to node[midway, color=blue, fill=white] {\begin{minipage}{3cm}\centering \small $V_3, V_6, V_7,$\\${{V_{10}}’}, {{V_{11}}’}, {{V_{12}}’},$\\${{V_{13}}’}$\end{minipage}} (11);
                \draw[thick] (11) -- (12) node[midway, color=blue,fill=white, swap] {\small$V_{7}, V'_{10}, V'_{11}, V'_{12}$};
                \draw[thick] (12) -- (13) node[midway, color=blue,fill=white, swap] {\small$V'_{11}, V'_{12}$};

             %   \draw[thick] (1) edge  (2);
             %  \draw[thick] (1) edge  (3);
             %   \draw[thick] (1) edge  (4);
              % \draw[thick] (1) edge[bend right=45]  (8);

             %\draw[thick] (2) edge  (5);
              % \draw[thick] (2) edge  (6);
               %     \draw[thick] (2) edge  (7);
               %     \draw[thick] (2) edge[bend right=25]  (8);    

               %   \draw[thick] (8) edge  (9);
               %     \draw[thick] (8) edge  (10);
               %     \draw[thick] (8) edge[bend left=35]  (11);                           

        \end{tikzpicture}}  
    \caption{}
    \label{fig:enter-label}
\end{figure}

\begin{table}[t]
       \centering
        \captionsetup{justification=centering}
        \caption{
        Results of empirical plug in for different sample sizes. Here $t$ is the tightness of the data, $k$ is the domain sizes of the variables, and recorded are the max table size during computation, and the sum of all table sizes.}
    \begin{subtable}{\columnwidth}
        \centering
        \caption{\scriptsize Cone Cloud $P(V_0 | do(V_{14}, V_{10}, V_{4}))$\\ $|\V| =15$; $ ~|\U|=8$; $ k=10$; hw $=2$; $tw=16$\\
        densest factor $P(V_0 | V_1 \ldots V_{13})$}
            \scriptsize    
        \resizebox{\columnwidth}{!}{\begin{tabular}{cccccc}
            \toprule
            \textbf{\#Samples} & \textbf{time} & \textbf{Max table size} & \textbf{Sum} & \textbf{t}& \textbf{density}\\
            \midrule
    100    & 2.2    & 9900     & 11825     & 100  & 1e-13 \\
    200    & 3.3    & 39203    & 43187     & 200  & 2e-13 \\
    400    & 7.3    & 152877   & 161573    & 400  & 4e-13 \\
    800    & 28.8   & 589815   & 612325    & 800  & 8e-13 \\
    1000   & 39.1   & 882604   & 913859    & 1000 & 1e-12 \\
    1600   & 84.8   & 2206528  & 2274829   & 1600 & 1.6e-12 \\
    3200   & 280.3  & 7553700  & 7816972   & 3200 & 3.2e-12 \\
    6400   & 788.74 & 21949125 & 23051073  & 6400 & 6.4e-12 \\
    10000  & 1594.2 & 40404630 & 43197340  & 10000 & 1e-11 \\
            \bottomrule
        \end{tabular}}
    \end{subtable}\\
     \begin{subtable}{\columnwidth}
        \centering
        \caption{\scriptsize Diamond Model, $P(V_{16}| do(V_0, V_4))$\\ $|\V| =17$;$k=10$; hw $=2$, tw=16\\$P(V_{16} |V_0, \ldots V_{15})$- densest factor}
            \scriptsize    
       \resizebox{\columnwidth}{!}{ \begin{tabular}{cccccc}
            \toprule
            \textbf{\#Samples} & \textbf{time} & \textbf{Max table size} & \textbf{Sum} & \textbf{t} &\textbf{density}\\
            \midrule
    100    & 0.6     & 8,672       & 22,265      & 100     & $1 \times 10^{-15}$  \\
200    & 1.9     & 35,605      & 81,922      & 200     & $2 \times 10^{-15}$  \\
400    & 6.9     & 156,160     & 334,874     & 400     & $4 \times 10^{-15}$  \\
800    & 26.4    & 603,790     & 1,254,919   & 800     & $8 \times 10^{-15}$  \\
1000   & 39      & 928,510     & 1,917,150   & 1000    & $1 \times 10^{-14}$  \\
1600   & 87.4    & 2,184,880   & 4,477,070   & 1600    & $1.6 \times 10^{-14}$\\
3200   & 299.4   & 7,600,350   & 15,547,839  & 3200    & $3.2 \times 10^{-14}$\\
6400   & 988.8   & 22,208,750  & 46,493,438  & 6400    & $6.4 \times 10^{-14}$\\
10,000 & 2037.5  & 39,943,750  & 87,542,864  & 10,000  & $1 \times 10^{-13}$  \\

        \bottomrule

\end{tabular}}
    \label{tab:3layer1_plug_in}
    \end{subtable}\\
    \shrink{
    \begin{subtable}{\columnwidth}
        \centering
        \caption{\scriptsize 2 Layer Model, $P(Z| do(T))$\\ $|\V| =12$;$|4 \leq  k \leq 50$; hw $=2$, tw=13\\$P(Z |V_0, \ldots V_{11})$- densest factor}
            \scriptsize    
       \resizebox{\columnwidth}{!}{ \begin{tabular}{cccccc}
            \toprule
            \textbf{\#Samples} & \textbf{time} & \textbf{Max table size} & \textbf{Sum} & \textbf{t} & \textbf{density}\\
            \midrule
100     & 1.3     & 18078     & 54444     & 98     & \( 1.5 \times 10^{-6} \) \\
200     & 2.8     & 45549     & 134503    & 195    & \( 2.9 \times 10^{-6} \) \\
400     & 5.7     & 91081     & 267559    & 378    & \( 5.6 \times 10^{-6} \) \\
800     & 9.2     & 130253    & 387764    & 699    & \( 1.0 \times 10^{-5} \) \\
1000    & 9.1     & 130944    & 401145    & 851    & \( 1.3 \times 10^{-5} \) \\
1600    & 12.4    & 166629    & 524332    & 1256   & \( 1.9 \times 10^{-5} \) \\
3200    & 17.7    & 237473    & 760792    & 2132   & \( 3.2 \times 10^{-5} \) \\
6400    & 27.3    & 280100    & 970221    & 3317   & \( 4.9 \times 10^{-5} \) \\
10,000  & 28.2    & 356559    & 1201240   & 4087   & \( 6.1 \times 10^{-5} \) \\
100,000 & 69.6    & 741134    & 2545997   & 9560   & \( 1.4 \times 10^{-4} \) \\
            \bottomrule
        \end{tabular}}
            \label{tab:cone_cloud_plug_in}

    \end{subtable}
    \begin{subtable}{\columnwidth}
        \centering
        \caption{\scriptsize 3 Layer Model, $P(Z| do(T))$\\ $|\V| =17$;$4 \leq  k \leq 20$; hw $=4$, tw=15\\$P(Z |V0, \ldots V_{11})$- densest factor}
            \scriptsize    
       \resizebox{\columnwidth}{!}{ \begin{tabular}{cccccc}
            \toprule
            \textbf{\#Samples} & \textbf{time} & \textbf{Max table size} & \textbf{Sum} & \textbf{t} &\textbf{density}\\
            \midrule

\end{tabular}}
    \label{tab:3layer1_plug_in}
    \end{subtable}}
    \begin{subtable}{\columnwidth}
        \centering
        \caption{\scriptsize 3 Layer Model, $P(\textbf{V}| do(Z,T))$\\ $|\V| =17$; $2 \leq k \leq 50$; hw $=4$, tw=15\\ $P(V_0 |V_1, V_2, V_3, X_0)$- densest factor }
            \scriptsize    
       \resizebox{\columnwidth}{!}{ \begin{tabular}{cccccc}
            \toprule
            \textbf{\#Samples} & \textbf{time} & \textbf{Max table size} & \textbf{Sum} & \textbf{t} & \textbf{density} \\
            \midrule
25   & 2.8     & 3,219      & 26,475      & 25  & 0.00125  \\
50   & 26.5    & 79,136     & 680,983     & 50  & 0.0025   \\
100  & 640.8   & 1,623,559  & 17,930,105  & 100 & 0.005    \\
200  & 30,511.5 & 34,018,439 & 363,988,664 & 200 & 0.01     \\
\end{tabular}}
    \label{tab:3layer2_plug_in}
    \end{subtable}
\begin{subtable}{\columnwidth}
    \caption{\scriptsize Chain: $P(V_{99} | do(V_{0}))$\\ $|\V| =99$; $ ~|\U|=49$ hw $=1$, tw=98\\ $P(V_{98} |V0, \ldots V_{97} )$- densest factor }
    \scriptsize    
      \resizebox{\columnwidth}{!}{%
    \begin{tabular}{  c c c c c c }%
    \toprule
    \textbf{\#Samples} & \textbf{time} & \textbf{Max table size} & \textbf{Sum} & \textbf{t} & \textbf{density} \\
     \midrule
  100 & 11.2 & 100 & 10109 & 100 & $9.9 \times 10^{-34}$ \\
200 & 19.7 & 200 & 20210 & 200 & $2.0 \times 10^{-33}$ \\
400 & 31.4 & 400 & 40410 & 400 & $3.9 \times 10^{-33}$ \\
800 & 53.7 & 800 & 80813 & 800 & $7.9 \times 10^{-33}$ \\
1000 & 69.5 & 1000 & 101014 & 1000 & $1.0 \times 10^{-32}$ \\
1600 & 97.5 & 1600 & 161612 & 1600 & $1.6 \times 10^{-32}$ \\
3200 & 192.4 & 3200 & 323214 & 3200 & $3.2 \times 10^{-32}$ \\
6400 & 369.2 & 6400 & 646415 & 6400 & $6.4 \times 10^{-32}$ \\
10000 & 608.8 & 10000 & 1010015 & 10000 & $9.9 \times 10^{-32}$ \\
    \bottomrule
    \end{tabular}}
    \label{tab:99chain_plug_in}
\end{subtable}\\
\label{tab:exp-results}
\end{table}
\clearpage
\begin{strip}
\begin{small}
\begin{multline}\label{eq:2layer}
P(Z | do(T)) = \sum_{V0,V1,V2,V3,V4,V5,V6,V7,V8,V9,V10,V11}P(Z|V0,V1,V2,V3,V4,V5,V6,V7,V8,V9,V10,V11)P(T)\\\sum_{X}P(X|T)P(V0|V1,V2,V3,X)P(V3|V4,V5,V6,X)P(V6|V7,V8,V9,X)P(V9|V10,V11,X)
\end{multline}
\end{small}
\end{strip}%
\begin{strip}
\begin{small}
\begin{multline}\label{eq:3layer_1}
P(Z \mid do(T)) = \sum\limits_{V_0, V_1, V_2, V_3, V_4, V_5, V_6, V_7, V_8, V_9, V_{10}, V_{11}} P(Z \mid V_0, V_1, V_2, V_3, V_4, V_5, V_6, V_7, V_8, V_9, V_{10}, V_{11}) P(T)\\
\sum\limits_{X_0, X_1, X_2} P(X_0 \mid T) P(X_1 \mid T) P(X_2 \mid T)P(V_0 \mid V_1, V_2, V_3, X_0) P(V_3 \mid V_4, V_5, V_6, X_0) P(V_6 \mid V_7, V_8, V_9, X_0) P(V_9 \mid V_{10}, V_{11}, X_0) \\P(V_1 \mid X_1) P(V_4 \mid X_1) P(V_7 \mid X_1) P(V_{10} \mid X_1) P(V_2 \mid X_2) P(V_5 \mid X_2) P(V_8 \mid X_2) P(V_{11} \mid X_2) 
\end{multline}
\end{small}
\end{strip}% 
\begin{strip}
\begin{small}
\begin{multline}\label{eq:3layer_2}
P(\textbf{V} \mid do(T,Z)) = \sum_{X_0, X_1, X_2} P(X_0 \mid T) P(X_1 \mid T) P(X_2 \mid T) P(T) 
 P(V_0 \mid V_1, V_2, V_3, X_0) P(V_3 \mid V_4, V_5, V_6, X_0)\\ P(V_6 \mid V_7, V_8, V_9, X_0) P(V_9 \mid V_{10}, V_{11}, X_0) P(V_1 \mid X_1)  P(V_4 \mid X_1) P(V_7 \mid X_1) P(V_{10} \mid X_1) P(V_2 \mid X_2) P(V_5 \mid X_2) \\
 P(V_8 \mid X_2) P(V_{11} \mid X_2) \hspace{50mm}
\end{multline}
\end{small}
\end{strip}% 

\begin{strip}
\begin{small}
\begin{multline}\label{eq:diamond}
P\left(V_{16} \middle| do(V_0), do(V_4)\right) = \sum_{V_5,V_6,V_7,V_8,V_9,V_{10},V_{11},V_{12},V_{13},V_{14},V_{15}}P\left(V_{15} \middle| V_0,V_{13},V_{14}\right) \times 
\frac{numerator}{denominator}
P\left(V_{13} \middle| V_0\right)P\left(V_{11} \middle| V_0,V_9,V_{10}\right)\\
\times P\left(V_9 \middle| V_0\right)P\left(V_7 \middle| V_0,V_5,V_6\right)P\left(V_5 \middle| V_0\right)
\\
\sum_{V_0'}P\left(V_{12} \middle| V_0',V_9,V_{10},V_{11}\right)
P\left(V_{10} \middle| V_0',V_9\right)P\left(V_0'\right)
\\
\sum_{V_0'}P\left(V_8 \middle| V_0',V_5,V_6,V_7\right)P\left(V_6 \middle| V_0',V_5\right)P\left(V_0'\right)
\end{multline}
\end{small}
\end{strip}% 
\begin{strip}
\begin{small}
\begin{multline}\label{eq:numerator}
numerator= \sum_{V_0',V_2,V_6,V_{10}}P\left(V_{16} \middle| V_0',V_1,V_2,V_3,V_4,V_5,V_6,V_7,V_8,V_9,V_{10},V_{11},V_{12},V_{13},V_{14},V_{15}\right)P\left(V_4,V_{14} \middle| V_0',V_1,V_2,V_3,V_5,V_6,V_7,V_8,V_9,V_{10},V_{11},V_{12},V_{13}\right)\\
P\left(V_2 \middle|V_0',V_1,V_5,V_6,V_7,V_8,V_9,V_{10},V_{11},V_{12},V_{13}\right)
P\left(V_8 \middle| V_0',V_5,V_6,V_7,V_9,V_{10},V_{11},V_{12},V_{13}\right)
P\left(V_6 \middle|V_0',V_5,V_9,V_{10},V_{11},V_{12},V_{13}\right)\\
P\left(V_{12} \middle| V_0',V_9,V_{10},V_{11},V_{13}\right)P\left(V_{10} \middle| V_0',V_9,V_{13}\right)P\left(V_0'\right)\\
\end{multline}
\end{small}
\end{strip}% 
\begin{strip}
\begin{small}
\begin{multline}\label{eq:denominator}
denominator=\sum_{V_0',V_2,V_6,V_{10}}{P\left(V_4 \middle| V_0',V_1,V_2,V_3,V_5,V_6,V_7,V_8,V_9,V_{10},V_{11},V_{12},V_{13}\right)P\left(V_2 \middle| V_0',V_1,V_5,V_6,V_7,V_8,V_9,V_{10},V_{11},V_{12},V_{13}\right)P\left(V_8 \middle| V_0',V_5,V_6,V_7,V_9,V_{10},V_{11},V_{12},V_{13}\right)P\left(V_6 \middle| V_0',V_5,V_9,V_{10},V_{11},V_{12},V_{13}\right)P\left(V_{12} \middle| V_0',V_9,V_{10},V_{11},V_{13}\right)P\left(V_{10} \middle| V_0',V_9,V_{13}\right)P\left(V_0'\right)}
\end{multline}
\end{small}
\end{strip}%
}

\end{document}